%% file: neurips_2026.tex
\documentclass{article}

    \PassOptionsToPackage{numbers, compress}{natbib}
 \usepackage[preprint]{neurips_2026}

\input{preamble.tex}

\input{math_commands.tex}

\usepackage[utf8]{inputenc} 
\usepackage[T1]{fontenc}    
\usepackage{hyperref}       
\usepackage{url}            
\usepackage{booktabs}       
\usepackage{amsfonts}       
\usepackage{nicefrac}       
\usepackage{microtype}      
\usepackage{xcolor}         

\title{Feedforward Novel View Synthesis for Heterogeneous Cameras}

\author{%
  Meng Wei$^{1}$ \quad Cheng Zhang$^{1}$ \quad Boying Li$^{1}$\thanks{Corresponding author} \quad Yihang Chen$^{3, 1}$ \\ \textbf{Jianmin Zheng$^2$ \quad Hamid Rezatofighi$^{1}$ \quad Jianfei Cai$^1$} \\
  $^1$Monash Univeristy \quad $^2$Nanyang Technological University \quad $^3$Shanghai Jiaotong University \\
  \texttt{\{meng.wei,cheng.zhang,boying.li\}@monash.edu} \\
  \texttt{\{yhchendut\}@gmail.com}  \quad  \texttt{\{ASJMZheng\}@ntu.edu.sg} \\
  \texttt{\{hamid.rezatofighi,jianfei.cai\}@monash.edu}
}

\begin{document}

\maketitle
\input{section/0_abstract.tex}
\input{section/1_intro.tex}
\input{section/2_related.tex}
\input{section/3_method.tex}

\input{section/4_experiment.tex}
\input{section/5_conclusion.tex}


\medskip

{
\small

\bibliographystyle{unsrtnat}
\bibliography{main, foundation}
}


\appendix
\input{section/x_supp.tex}


\newpage
\input{checklist.tex}

\end{document}

%% file: preamble.tex
\usepackage{times}
\usepackage{epsfig}
\usepackage{graphicx}
\usepackage{amsmath}
\usepackage{amssymb}
\usepackage{colortbl}
\usepackage{comment}
\usepackage{caption}
\usepackage{subcaption}
\usepackage{multirow}
\usepackage{wrapfig}

\usepackage[utf8]{inputenc} 
\usepackage[T1]{fontenc}    
\usepackage{url}            
\usepackage{booktabs}       
\usepackage{amsfonts}       
\usepackage{nicefrac}       
\usepackage{microtype}      
\usepackage{xcolor}         
\usepackage{pifont}         
\usepackage{hyperref}       

\definecolor{best}{rgb}{0.96, 0.57, 0.58}
\definecolor{second}{rgb}{0.98, 0.78, 0.57}
\definecolor{third}{rgb}{1.0, 1.0, 0.56}

\definecolor{darkpink}{rgb}{0.91, 0.33, 0.51} 
\definecolor{deepmagenta}{RGB}{219, 48, 122} 

%% file: math_commands.tex
\usepackage{amsmath,amsfonts,bm}

\def\eqref#1{equation~\ref{#1}}

\def\1{\bm{1}}

\DeclareMathAlphabet{\mathsfit}{\encodingdefault}{\sfdefault}{m}{sl}
\SetMathAlphabet{\mathsfit}{bold}{\encodingdefault}{\sfdefault}{bx}{n}



%% file: section/0_abstract.tex
\begin{abstract}
Feed-forward novel view synthesis has recently shown promising results from sparse posed images, but most existing methods assume that context and target views share a fixed camera family. This homogeneous-camera assumption breaks in practical multi-sensor systems, where perspective, fisheye, and panoramic cameras may coexist and where the target projection may be unseen during training. We study feed-forward NVS across heterogeneous central cameras and identify a key ambiguity introduced by tokenization: a visual token aggregates a projection-dependent bundle of pixel rays, while existing camera encodings mainly expose absolute rays or token-center relations. To address this, we combine token-center relative Camera Positional Encodings and proposed local raymaps, a token-level representation that explicitly describes the intra-patch ray distribution summarized by each token. We further propose projection-aware 2D RoPE, which replaces raw image-grid coordinates with ray-induced angular coordinates so that relative positional reasoning is aligned across camera projections. Together, these components treat diverse cameras as calibrated samplings of a shared ray space rather than separate visual domains. On ScanNet++ with heterogeneous-camera system, our method improves over camera-conditioned baselines under mixed-camera evaluation and demonstrates zero-shot generalization to panoramic views.
\end{abstract}

%% file: section/1_intro.tex
\section{Introduction}\label{sec:intro}
Novel view synthesis (NVS) aims to synthesize unseen views from posed observations and remains a central topic in robotics, embodied perception, and world models. Recent feed-forward NVS methods have made substantial progress by predicting novel views from sparse posed images in a single forward pass~\cite{yu2021pixelnerf,chen2021mvsnerf,charatan2023pixelsplat,chen2024mvsplat,jin2025lvsm}. Most of these methods, however, are designed, trained, and evaluated under a homogeneous-camera assumption: the input and target views share a fixed projection family, such as perspective~\cite{jin2025lvsm} or panoramic cameras~\cite{zhang2024pansplat}, but with mutually exclusive camera-specific architectures.

This assumption excludes many practical capture and rendering systems. Multi-sensor platforms often combine perspective, fisheye, and panoramic cameras, while downstream applications may require rendering into a camera geometry different from the input sensors. In this setting, NVS is not only a problem of interpolating views in 3D space, but also one of reasoning about visual evidence across different samplings of ray space. A basic modeling question follows: \emph{can a single feed-forward model support heterogeneous cameras, even for camera types unseen during training?}

A purely data-driven approach would expose the model to different intrinsics, distortions, and cross-camera capture conditions. However, data-driven strategy is difficult to scale because camera variation is continuous and real multi-camera data introduce non-geometric factors such as exposure, white balance, and calibration noise. More importantly, heterogeneous inference requires the model to exhibit \emph{few- or zero-shot generalizability} to camera types broader than training data. A reusable geometric principle for unseen projections is preferable to exhaustive memorization.

This motivates a local view of camera generalization. Although global projection functions can differ substantially, their local ray configurations recur across camera types. An unseen camera may define a new full-image projection, while many of its patch-level ray patterns remain geometrically similar to local patterns observed during training. We therefore formulate heterogeneous-camera generalization as learning over reusable local ray geometry, rather than over discrete camera categories or exhaustively sampled camera instances.

Existing camera encodings, in contrast, operate at three other granularities: absolute pixel-level rays in Pl\"ucker raymaps~\cite{jin2025lvsm}, relative camera geometry between views~\cite{li2025prope,Miyato2024GTA}, and relative token-center rays~\cite{wu2026rayrope,zhang2025ucpe}. None of them describes the intra-patch ray bundle each token aggregates: two tokens with the same center ray can still cover very different ray distributions under different projections, leaving the patch-level ray geometry implicit, resulting in unnecessary learning burdens.

Based on these observations, we introduce local raymaps for feed-forward NVS across heterogeneous central cameras, complementary to token-center relative Camera Positional Encodings~\cite{zhang2025ucpe}. A local raymap represents the patch-level ray configuration associated with each visual token, exposing the intra-patch ray geometry that is otherwise hidden by tokenization. We further propose projection-aware 2D RoPE~\cite{su2024roformer} in ray-induced angular coordinates to align relative positional reasoning across camera projections. Under this derivation, perspective, fisheye, and equirectangular views are treated as calibrated samplings of a shared ray space rather than as separate visual domains. We evaluate the proposed formulation on ScanNet++ v2~\cite{yeshwanthliu2023scannetpp} by constructing heterogeneous-camera NVS tasks from undistorted, fisheye, and panoramic images, including mixed-camera evaluation and zero-shot target-camera evaluation.

In summary, our contributions are:
\begin{itemize}
    \item We introduce heterogeneous-camera feed-forward NVS, a setting where context and target views may use different central projection models, and define mixed-camera and zero-shot target-camera evaluation protocols on ScanNet++.
    \item We apply a token-center relative Camera Positional Encodings and propose local raymaps, a token-level representation that exposes reusable intra-patch ray geometry without relying on discrete camera categories, which exhibits zero-shot generalizability, supporting heterogeneous-camera NVS.
    \item We introduce projection-aware 2D RoPE in ray-induced angular coordinates, aligning relative positional reasoning across camera projections.
    \item We evaluate heterogeneous-camera NVS on ScanNet++ under mixed-camera and zero-shot target-camera protocols and achieve the state-of-the-art performance.
\end{itemize}

%% file: section/2_related.tex
\section{Related Work}
\paragraph{Wide-FoV Cameras Across 3D Tasks.}
Wide-FoV cameras, including fisheye and panoramic cameras, are widely used in embodied perception and AR/VR because they observe much larger scene extents than standard perspective cameras. This has motivated projection-aware modeling across several tasks. For reconstruction, panoramic and universal-camera depth methods estimate dense geometry from equirectangular, fisheye, or arbitrary camera inputs~\cite{cao2025panda,guo2025depth,piccinelli2025unik3d}, while feed-forward 3D reconstruction models extend point-map or point-cloud prediction to panoramic and fisheye images~\cite{guo2026panovggt,duan2026fisheye3r}. For NVS and radiance-field rendering, omnidirectional and fisheye Gaussian-splatting methods adapt optimization and rasterization to non-perspective projection rays~\cite{lee2024odgs,li2025omnigs,liao2024fisheye,wu20253dgut}. Wide-FoV images have also become a common interface for video generation and 3D world modeling through panorama diffusion and world-generation methods~\cite{wang2024360dvd,zhang2026panflow,tan2024imagine360,luo2025beyond,yang2025matrix,wang2025evoworld,team2025hunyuanworld}. Together, these works show that wide-FoV images require geometry beyond ordinary image coordinates, motivating projection-aware novel view synthesis.

\paragraph{Feed-forward Novel View Synthesis.}
Novel view synthesis was popularized in recent learning-based 3D vision through per-scene radiance field optimization, including NeRF~\cite{mildenhall2021nerf} and its successors~\cite{3DGS_2023,barron2023zipnerf}.
Feed-forward NVS~\cite{yu2021pixelnerf,chen2021mvsnerf,charatan2023pixelsplat,chen2024mvsplat,chen2024mvsplat360,zhang2024pansplat,jin2025lvsm} instead synthesizes novel views, or predicts an intermediate 3D representation, directly from input images. Most methods are developed for standard perspective cameras: early generalizable radiance-field methods achieve neural rendering or predict 3D Gaussians using learned image features, while recent Transformer-based models reduce explicit 3D inductive bias and learn to directly synthesize target images, as in LVSM~\cite{jin2025lvsm}.

Recent work extends feed-forward NVS toward wide-FoV cameras, with PanSplat and Splatter-360 introducing specialized designs for sparse-view or feed-forward panorama synthesis~\cite{zhang2024pansplat,chen2024splatter360}. These camera-specific systems are effective, but both perspective and wide-FoV NVS methods usually assume a fixed camera family during training and evaluation. This leaves open a more general question: can a feed-forward NVS model reason when context and target views are produced by different projection models, possibly including target camera types unseen during training? We study this heterogeneous-camera setting by treating perspective, fisheye, and panoramic cameras as calibrated ray samplings of the same scene, and ask what information allows one model to share scene reasoning across them.

\paragraph{Camera Positional Encodings.}
Reasoning across such cameras requires exposing camera geometry to the model. Transformers require positional information to make token interactions spatially meaningful~\cite{attention,dosovitskiy2020vit,liu2021swinTrans}; for multi-view generation and reconstruction, this position is not only a 2D image coordinate but also the camera geometry that relates tokens across views. Existing methods therefore encode cameras either as token content or as attention positional encodings. LVSM provides absolute Pl\"ucker raymaps as token content, allowing context images and target cameras to be represented in a common 3D frame~\cite{jin2025lvsm}. PRoPE formulates cameras as relative positional encodings, injecting camera pose and intrinsics into attention so that query-key interactions depend on projective geometry~\cite{li2025prope}. RayRoPE further studies ray-based positional encoding for multi-view attention~\cite{wu2026rayrope}, and UCPE expresses camera geometry through calibrated rays to support different projection models and lens distortions~\cite{zhang2025ucpe}.

Despite these advances, camera positional encodings alone do not remove the ambiguity of token ray distributions, which is a key obstacle for reasoning across heterogeneous cameras. Attention-level encodings such as RayRoPE and UCPE describe geometric relations between token centers, but each patch token aggregates multiple pixel rays whose local distribution depends on camera projection. Our local raymaps are therefore complementary: they provide token-level evidence about the intra-patch ray bundle, while existing camera positional encodings provide cross-token or cross-view relations.

%% file: section/3_method.tex
\section{Method}\label{sec:method}
We study feed-forward novel view synthesis for heterogeneous central cameras, where context and target views may come from pinhole, fisheye, or panoramic cameras. The key difficulty is not merely representing multiple camera types: calibrated rays already provide a common interface for token inputs and camera positional encodings, as reviewed in Sec.~\ref{sec:lvsm}. Instead, heterogeneity changes the geometric meaning of Transformer tokens themselves. After patchification, each token summarizes a projection-dependent ray distribution, and the image lattice no longer provides a camera-invariant coordinate system for attention.

To support heterogeneous NVS, we first extend the token content with \emph{local raymaps} (Sec.~\ref{sec:local_raymaps}), allowing the model to reuse local ray patterns rather than infer camera categories from data coverage. We then replace raw image coordinates in 2D RoPE with ray-induced angular coordinates (Sec.~\ref{sec:proj_rope}), aligning relative positional phases across projection models. These components allow a single decoder-only Transformer to operate across mixed and even unseen camera geometries. 
Fig.~\ref{fig:arch} summarizes how our encoding exposes projection-consistent token geometry compared with prior camera encodings.

\input{pics/method_arch.tex}

\subsection{Prerequisites}
\paragraph{Problem setup.}
Let $\mathcal{C}=\{(I_i,\Pi_i,\mathbf{T}_i^{\mathrm{cw}})\}_{i=1}^{N}$ denote context images $I_i$ with known projection models $\Pi_i$ and world-to-camera poses $\mathbf{T}_i^{\mathrm{cw}}$. We define each projection model as a pixel-to-ray mapping $\Pi_i:\Omega_i\rightarrow\mathbb{S}^2$, which maps an image coordinate $\mathbf{u}\in\Omega_i$ to a unit ray direction $\Pi_i(\mathbf{u})$ in the local camera frame. Given a target camera $(\Pi_t,\mathbf{T}_t^{\mathrm{cw}})$, feed-forward NVS predicts the target image $\tilde{I}_t$ in a single forward pass. We study the heterogeneous-camera setting, where the projection models $\Pi_i$ and $\Pi_t$ may differ across views and may even include camera types unseen during training. We assume central cameras, where all camera rays within one image pass through a single camera center.

\paragraph{Transformer-based NVS.}\label{sec:lvsm}
Our backbone follows the decoder-only Transformer formulation of LVSM~\cite{jin2025lvsm}. The model converts context observations and the target camera into a single sequence of patch tokens~\cite{dosovitskiy2020vit}, applies self-attention, and decodes the output tokens at target positions into $\tilde{I}_t$.

In LVSM, camera geometry is provided as an absolute Pl\"ucker raymap $R_i^{\mathrm{pl}}$, which stores the 3D ray associated with each image location in a common coordinate frame. For a context view, the RGB image and its raymap are tokenized jointly; for the target view, only the target raymap is tokenized:
\begin{equation}
    \mathbf{z}_{i,p} = \theta_{\mathrm{ctx}}\!\left(I_i, R_i^{\mathrm{pl}}\right)_p,
    \qquad
    \mathbf{z}_{t,p} = \theta_{\mathrm{tgt}}\!\left(R_t^{\mathrm{pl}}\right)_p ,
\end{equation}
where $p$ indexes image patches, and $\theta_{\mathrm{ctx}}$ and $\theta_{\mathrm{tgt}}$ denote the context and target tokenizers. The resulting context tokens $\{\mathbf{z}_{i,p}\}$ and target tokens $\{\mathbf{z}_{t,p}\}$ are concatenated before entering the Transformer.
This formulation makes pixel-aligned camera rays available as token content, while geometric relations between tokens are still learned implicitly through attention.

\paragraph{Cameras as positional encodings.}\label{sec:cameraPE}
Camera positional encodings inject camera geometries directly into attention operations, complementary to input raymaps.
Following RoPE-style formulations~\cite{su2024roformer}, each token $p$ is assigned a camera-aware transformation matrix $\mathbf{D}_p$:
\begin{equation}
    \mathbf{D}_p = \mathbf{I}_{d/4} \otimes \mathbf{T}_{p},
\end{equation}
where $\mathbf{I}$ is the identity matrix, $d$ is the feature dimension, $\otimes$ denotes the Kronecker product, and $\mathbf{T}_{p}$ specifies the camera-dependent transform for token $p$.
During attention, the query and key features are transformed by $\mathbf{D}$ before computing attention scores:
\begin{equation}
    O = \mathrm{Attn}\!\left(\mathbf{D}^\top Q, \mathbf{D}^{-1} K, V\right).
\end{equation}
For a query token $q$ and a key token $k$, the attention logit is therefore modulated by the relative transform between their geometries:
\begin{equation}
    \left(\mathbf{D}_q^\top Q_q\right)^\top
    \left(\mathbf{D}_k^{-1} K_k\right)
    =
    Q_q^\top \mathbf{D}_q\mathbf{D}_k^{-1} K_k,
    \qquad
    \mathbf{D}_q\mathbf{D}_k^{-1}
    =
    \mathbf{I}_{d/4}\otimes \mathbf{T}_{q}\mathbf{T}_{k}^{-1}.
\end{equation}

The definition of $\mathbf{T}_p$ therefore determines which geometric relation is visible to attention. CaPE~\cite{Xiong2023CAPE} uses the camera pose $\mathbf{T}_{c(p)}^{\mathrm{cw}}$, where $c(p)$ is the camera index associated with token $p$. 
Geometric Transform Attention (GTA)~\cite{Miyato2024GTA} additionally transforms the value matrix. Projective Positional Encoding (PRoPE)~\cite{li2025prope} incorporates camera intrinsics $\mathbf{K}$ and defines the camera-frame raymaps, named CamRay. RayRoPE~\cite{wu2026rayrope} and Unified Camera Positional Encoding (UCPE)~\cite{zhang2025ucpe} define positional relations using calibrated ray geometry, extending attention-level encodings to diverse camera models and lens distortions. 
Our method thus builds on the ray-based UCPE formulation, which defines $\mathbf{T}_p$ as a function of the token-center ray direction, compatible with diverse camera models.

\subsection{Extending Camera Positional Encodings for Heterogeneous Cameras}\label{sec:local_raymaps}
The prerequisites above describe two common ways to expose camera geometry to a Transformer: raymaps provide token content, and camera positional encodings modulate attention through token-center geometry. These mechanisms are often sufficient in a homogeneous-camera setting because the image lattice carries a stable geometric meaning. When all views follow the same projection family, a patch thus corresponds to a predictable local angular footprint, and a displacement on the token grid is also a stable proxy for a displacement in ray space.

Heterogeneous cameras break this structural shortcut, creating a representation ambiguity for current Transformer-based methods, where tokenized visual content doesn't share similar geometric meanings. Tokens covering the same image-space pixel range may aggregate different ray bundles under different camera projections: it can cover a narrow perspective patch, a stretched fisheye content, or a panorama footprint near a different latitude. This mismatch can hinder the model from generalizing spatial relationships consistently across camera types.

To make this mismatch explicit, let $\Omega_p$ denote the pixel set of patch $p$, and let $\bar{\mathbf{u}}_p$ denote its center pixel. A context token is produced by jointly tokenizing the RGB patch and its camera representation,
\begin{equation}
    \mathbf{z}_{i,p}
    =
    \theta_{\mathrm{ctx}}\!\left(I_i, R_i\right)_p ,
\end{equation}
where $R_i$ denotes any camera ray representation used by the backbone, as in Fig.~\ref{fig:arch}. Since $\Pi_i$ maps pixels to camera-frame unit rays, the token-center ray is $\bar{\mathbf{r}}_{i,p} = \Pi_i(\bar{\mathbf{u}}_p),$ which is the primary geometric cue directly exposed to attention-level camera encodings.

However, the RGB evidence summarized by $\mathbf z_{i,p}$ is aggregated from the entire patch rather than a single pixel. Each token actually corresponds to a bundle of rays, whose support is defined as
\begin{equation}
    \mathcal{B}_{i,p}
    =
    \left\{
    \Pi_i(\mathbf{u})
    \mid
    \mathbf{u}\in\Omega_p
    \right\}.
\end{equation}
Overall, $\bar{\mathbf r}_{i,p}$ only captures the center viewing direction of the token, whereas $\mathcal B_{i,p}$ represents the entire local ray distributions over the whole patch. In homogeneous-camera NVS, the former serves as a compact surrogate for the latter because the projection model consistently determines how the local ray bundle varies around the center ray.
In heterogeneous-camera NVS, however, this surrogate relationship no longer holds reliably. For example, two key tokens, with indices $k_1, k_2$ and center-rays $\bar{\mathbf{r}}_{k_1}, \bar{\mathbf{r}}_{k_2}$, may share the same center-level geometric relation to a query token $q$ with the center-ray $\bar{\mathbf{r}}_{q}$, while aggregating different local ray distributions $\mathcal{B}$:
\begin{equation}
    g\!\left(\bar{\mathbf{r}}_{q},\bar{\mathbf{r}}_{k_1}\right)
    =
    g\!\left(\bar{\mathbf{r}}_{q},\bar{\mathbf{r}}_{k_2}\right),
    \quad
    \mathcal{B}_{k_1} \neq \mathcal{B}_{k_2},
\end{equation}
where $g(\cdot,\cdot)$ denotes the token-level relative ray relation visible to attention-level camera encodings. In this case, attention sees equivalent center-ray geometry even though the two keys aggregate different projection-induced ray bundles. The missing distinguishable variable is thus the local mapping from patch pixels to nearby rays.

\paragraph{From raymaps to local raymaps.}
Explicit raymaps provide a unified interface for heterogeneous cameras by specifying where sampled pixels look, potentially indicating the missing local ray statistics. However, world-space Pl\"ucker raymaps~\cite{jin2025lvsm} or camera-frame raymaps (CamRay~\cite{li2025prope}) inputs primarily encode absolute or center-aligned ray content, and therefore do not explicitly identify the local ray distributions hidden by patchification. This limitation may lead to the attention-sink behavior illustrated in Fig.~\ref{fig:attnSink}, where center-ray inputs such as CamRay encourage tokens to first attend to some fixed model-learned tokens, even when their locations mismatch.

Moreover, the absolute or center-aligned ray encodings introduce learning  burden for models to infer patch-scale angular extent from either RGB appearance or absolute ray cues. It thus requires diverse training coverage, limiting the generalization capabilities for inference with heterogeneous cameras.

\input{pics/attnSink.tex}

We address these issues by introducing \emph{local raymaps}, which explicitly describe the local ray distribution within each token, easing the learning burden. The local raymaps shift the generalization unit from global projection types to reusable local ray patterns. While full-image projection models may differ widely, their patch-level ray configurations can recur across camera systems, allowing the representation to transfer to unseen projections without relying on camera-categorical encodings.

\paragraph{Construction of local raymaps.}
We encode each local ray bundle in a coordinate system normalized by the patch center ray. Let $\bar{\mathbf{r}}_{i,p}=\Pi_i(\bar{\mathbf{u}}_p)$ be the ray through the center of patch $p$ in view $i$. For each patch, we construct a rotation $\mathbf{Q}_{i,p}\in \mathrm{SO}(3)$ that maps $\bar{\mathbf{r}}_{i,p}$ to the canonical forward direction $\mathbf{e}_z=(0,0,1)^\top$. In other words, its local $z$-axis is the center ray and other in-plane axes are fixed by a deterministic reference axis. In practice, we use the up vector to determine the roll.

We then use the constructed rotation $\mathbf{Q}_{i,p}$ to transform every ray in the patch, producing a locally normalized ray bundle.
Applying this normalization to all patches gives the local raymap $R^{\mathrm{local}}_i$.
We provide local raymaps as token-level geometric content together with the existing image:
\begin{equation}
    \mathbf{z}_{i,p}
    =
    \theta_{\mathrm{ctx}}\!\left(I_i,R^{\mathrm{local}}_i\right)_p,
    \qquad
    \mathbf{z}_{t,p}
    =
    \theta_{\mathrm{tgt}}\!\left(R^{\mathrm{local}}_t\right)_p ,
\end{equation}
where $R^{\mathrm{local}}_i$ and $R^{\mathrm{local}}_t$ denote the local raymaps for context and target views respectively.

This design is compatible with the token-center relative camera positional encodings, UCPE specifically. Center-level encodings guide which tokens should interact geometrically, while local raymaps describe what ray footprint each interacting token summarizes. As illustrated by the attention visualization in Fig.~\ref{fig:attnSink}, making the local footprint explicit reduces the burden on attention to recover patch-scale ray statistics from global or center-only ray cues, resulting in cleaner attention maps.

\subsection{Projection-Aware 2D RoPE for Heterogeneous Cameras}\label{sec:proj_rope}
Transformer-based feed-forward novel view synthesis requires a positional mechanism that makes token interactions aware of 2D spatial layout.
Recent Camera Positional Encodings follow the GTA~\cite{Miyato2024GTA} formulation which defines one SE(3) transform for camera geometries and another SO(2) transform (2D RoPE~\cite{su2024roformer}) for multi-frequency image-space positions, where 2D RoPE encodes relative positions on the raw image patch indices.

For homogeneous cameras, the patch indices have stable geometric meanings. A displacement of one patch horizontally or vertically matches a fixed displacement in the underlying image projection. Standard 2D RoPE is thus aligned with the camera model. Let $\mathbf{u}_p=(u_p,v_p)^\top$ denote the 2D patch coordinate of token $p$. 2D RoPE applies a coordinate-dependent rotation to queries and keys,
\begin{equation}
    \tilde{\mathbf{q}}_q = \mathbf{D}_{\textrm{RoPE}}(\mathbf{u}_q)\mathbf{q}_q,
    \qquad
    \tilde{\mathbf{k}}_k = \mathbf{D}_{\textrm{RoPE}}(\mathbf{u}_k)\mathbf{k}_k,
\end{equation}
where $\mathbf{D}_{\textrm{RoPE}}(\cdot)$ is a block-diagonal matrix of frequency-dependent rotations, which is applied to $u-$ and $v$-axis separately. The resulting attention logit depends on the relative lattice displacement,
\begin{equation}
    \tilde{\mathbf{q}}_q^\top \tilde{\mathbf{k}}_k
    =
    \mathbf{q}_q^\top
    \mathbf{D}_{\textrm{RoPE}}(\mathbf{u}_k-\mathbf{u}_q)
    \mathbf{k}_k .
\end{equation}
This formulation implicitly assumes that $\mathbf{u}_k-\mathbf{u}_q$ has a consistent geometric meaning across images. However, this assumption breaks under heterogeneous cameras. As illustrated by Fig.~\ref{fig:2DRoPE}, the same lattice displacement can correspond to different angular displacements, since the projection topology could change the mapping from image coordinates to rays. Thus, the raw image indices are no longer a camera-invariant positional domain for 2D RoPE.

We therefore replace raw image patch indices with ray-induced angular coordinates. Since our input is already conditioned on camera raymaps, we directly compute the RoPE coordinates from the ray direction of each token. Let $c_i$ denote the camera associated with token $i$, and let $\mathbf{d}_i^{c_i}=(d_{i,x}^{c_i},d_{i,y}^{c_i},d_{i,z}^{c_i})$ denote the unit ray direction of token $i$ in that camera's local frame. We then map each ray to a longitude-latitude coordinate,
\begin{equation}
    \boldsymbol{\eta}_i^{c_i}
    =
    h(\mathbf{d}_i^{c_i})
    =
    (\lambda_i^{c_i},\varphi_i^{c_i})^\top
    \in \mathbb{R}^2 ,
\end{equation}
where
\begin{equation}
    \lambda_i^{c_i}
    =
    \operatorname{atan2}(d_{i,x}^{c_i},d_{i,z}^{c_i}),
    \qquad
    \varphi_i^{c_i}
    =
    \operatorname{atan2}
    \left(
    d_{i,y}^{c_i},
    \sqrt{(d_{i,x}^{c_i})^2+(d_{i,z}^{c_i})^2}
    \right).
\end{equation}
Equivalently, since $\mathbf{d}_i^{c_i}$ is unit-normalized, $\varphi_i^{c_i}=\arcsin(d_{i,y}^{c_i})$. This coordinate is shared across camera models: pinhole, fisheye, and panoramic images may have different image lattices, but their tokens can all be mapped to consistent directions on the viewing sphere.

The RoPE rotations are then applied using the angular ray coordinate $\boldsymbol{\eta}_i^{c_i}$ rather than the raw patch coordinate:
\begin{equation}
    \tilde{\mathbf{q}}_q
    =
    \mathbf{D}_{\textrm{RoPE}}(\boldsymbol{\eta}_q^{c_q})\mathbf{q}_q,
    \qquad
    \tilde{\mathbf{k}}_k
    =
    \mathbf{D}_{\textrm{RoPE}}(\boldsymbol{\eta}_k^{c_k})\mathbf{k}_k .
\end{equation}
Thus, the relative phase is determined by angular displacement in the ray domain rather than by displacement on the raw image lattice. This preserves the relative-position structure of RoPE while making its coordinate domain consistent across heterogeneous camera models.

%% file: pics/method_arch.tex
\begin{figure}[t]
    \centering
    \includegraphics[width=1.\textwidth]{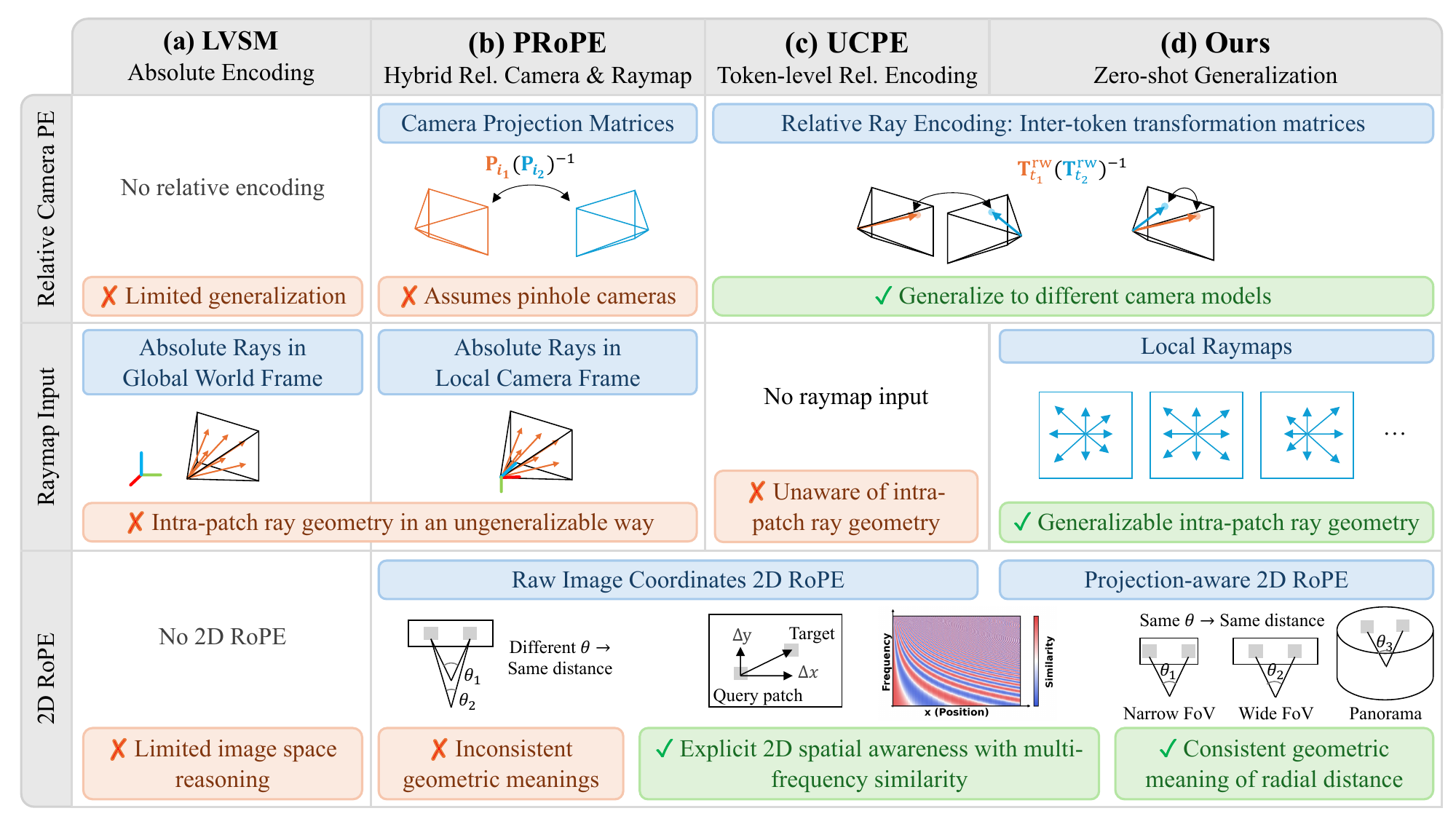}
    \caption{\textbf{Comparison of camera encoding methods.}
Existing methods encode absolute rays, center-level relations, or raw image coordinates, leaving projection-dependent token ray statistics implicit. Our method combines relative camera encoding, local raymaps, and projection-aware 2D RoPE to preserve consistent geometry across heterogeneous and unseen cameras.}
    \label{fig:arch}
\end{figure}

%% file: pics/attnSink.tex

\begin{figure}
    \centering
    \begin{minipage}{.65\textwidth}
        \centering
        \includegraphics[width=0.75\textwidth]{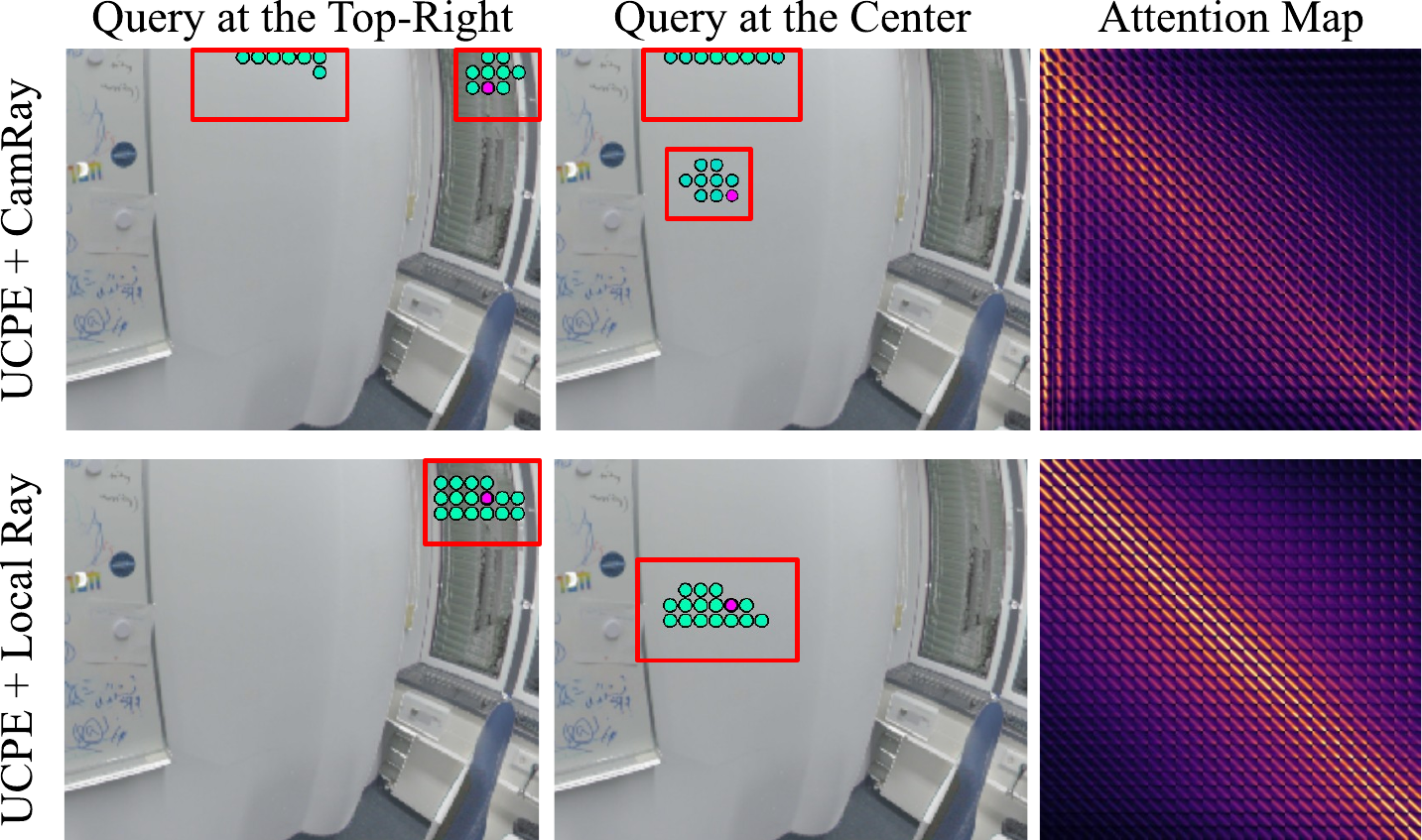}
        \captionof{figure}{\textbf{Attention sink phenomenon} in visualizations of the ``attended tokens with Top-16 weights'' (\textcolor{green!70!black}{green}) and attention maps (target view only) in the 1st layer, with target-view tokens as queries (\textcolor{darkpink}{pink}). Using CamRay representations makes tokens attend to learned \textbf{placeholder} tokens rather than to geometrically near tokens, while using local raymaps yields a cleaner locality.
        }
        \label{fig:attnSink}
    \end{minipage}
    \hfill
    \begin{minipage}{.3\textwidth}
        \centering
        \includegraphics[width=0.6\textwidth]{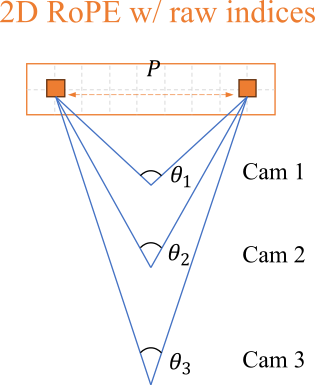}
        \captionof{figure}{\textbf{2D RoPE with raw image indices.} Across heterogeneous cameras, relative positions in the raw image indices may have inconsistent geometric meanings.}
        \label{fig:2DRoPE}
    \end{minipage}
    \vspace{-10pt}
\end{figure}

%% file: section/4_experiment.tex
\section{Experiments}\label{sec:experiments}
We evaluate the proposed method under heterogeneous-camera novel view synthesis, where accurate rendering requires both cross-view geometric reasoning and robustness to changes in camera projection. 
Our experiments study two complementary questions: whether the model improves reconstruction when camera types are mixed within the training distribution, and whether the learned representation transfers to a target camera type that is absent during training.

\subsection{Experimental Setup}
\paragraph{Dataset.}
We train and evaluate on ScanNet++ v2~\cite{yeshwanthliu2023scannetpp}, which captured indoor scenes with different cameras. 
We use fisheye (DSLR), undistorted, and panoramic images to construct heterogeneous-camera novel view synthesis pairs.
For the zero-shot setting, panoramic views are excluded during training and used only at test time, allowing us to evaluate generalization to unseen, challenging camera geometry.
We use the official training and validation splits, and use 224$\times$336 resolution for perspective cameras and 256$\times$512 for panorama cameras, at a similar scale to the training resolution used in PRoPE~\cite{li2025prope}.


\paragraph{Baselines.}
We compare our method with representative camera-conditioning strategies under heterogeneous camera settings. 
For token-level camera embeddings, we evaluate (1) \emph{the Pl\"{u}cker raymaps} used in LVSM~\cite{jin2025lvsm} and (2) \emph{the CamRay raymaps} introduced in PRoPE~\cite{li2025prope}. These baselines expose camera geometry directly to visual tokens, but differ in their coordinate dependence: Pl\"{u}cker raymaps encode rays in a global frame, whereas CamRay encodes local camera-frame ray directions. For attention-level camera positional encodings, we compare against PRoPE~\cite{li2025prope}, which injects relative projective camera geometry into self-attention, and UCPE~\cite{zhang2025ucpe}, which provides a ray-based relative camera encoding for diverse cameras. Our variants use UCPE as the camera positional encoding and differ in the token-level raymap and the coordinate domain of 2D RoPE.

\paragraph{Training details.}
All methods are trained in the same codebase with identical hyperparameters. 
Following PRoPE~\cite{li2025prope}, we adopt the lightweight LVSM~\cite{jin2025lvsm} backbone with approximately 24.5M parameters to match a practical academic compute budget. 
Each training sample contains two context views and one target view, following the default PRoPE setting. 
Unless otherwise specified, models are trained for 80k iterations on two NVIDIA A100 GPUs with a total batch size of 8. 
Additional implementation details are provided in the supplementary material.
We evaluate the performance of novel view synthesis using Peak Signal-to-Noise Ratio (PSNR), Structural Similarity Index Measure (SSIM~\cite{Wang2004SSIM}), and Learned Perceptual Image Patch Similarity (LPIPS~\cite{zhang2018lpips}) metrics.

\subsection{Evaluations}
\paragraph{In-distribution heterogeneous evaluation.}
We first evaluate NVS performance on fisheye and undistorted camera types. The evaluation protocols include two homogeneous-camera cases, ``fisheye $\to$ fisheye'' and ``undistorted $\to$ undistorted'', and a mixed-camera case where context and target views are randomly drawn from the available camera types.  We also apply the random focal length similar to PRoPE~\cite{li2025prope}. We use the Splatt3R benchmark~\cite{smart2024splatt3r} for view selection, which measures the overlapping ratio between the context views and the target view. While this overlapping ratio is measured on the undistorted images, we use the same camera poses for fisheye images as proximity.

\input{table/main.tex}

Table~\ref{tab:main} shows that relative ray-based camera positional encoding is an effective baseline for heterogeneous NVS, and that the proposed local raymaps further improve performance. Our methods achieve the best performance across all three protocols. Replacing raw 2D RoPE with projection-aware RoPE gives consistent additional gains over Ours (Raw), especially for fisheye and mixed-camera evaluation, where the image lattice is less geometrically stable. These results support the role of local ray distributions and projection-aware positional coordinates as complementary components.

\paragraph{Zero-shot generalizability.}
We next evaluate zero-shot generalization to panoramic target cameras, which are not used during training. We consider two protocols. In ``Virtual Cams $\rightarrow$ Panorama'', 6 context images are fisheye views synthesized from panoramas with random virtual camera models. In ``DSLR $\rightarrow$ Panorama'', context images come from real DSLR captures (fisheye) and targets are panoramas. To ensure enough coverage, we use 10 context images as inputs. The details of sampling method could be found in the supplementary materials. For all methods, we use the relative center-ray positional encoding (UCPE~\cite{zhang2025ucpe}). Since this setting involves cross-device appearance differences, we apply the same local histogram matching color correction module to all methods.
\input{table/pano.tex}
\input{pics/main.tex}
\input{pics/pano.tex}

Table~\ref{tab:pano_zeroshot} shows that local raymaps transfer better to unseen panorama geometry than global or center-only ray representations. CamRay fails in this zero-shot setting, suggesting that center-aligned ray directions alone are not sufficient to cover the unseen projection regime.

\subsection{Ablations}
There are two main components of our methods: the local raymap input and the projection-aware RoPE.
Previous experiments have shown the isolated effects of these components, where ``Ours (Raw)'' does not use the projection-aware RoPE and has shown performance gain over other raymap inputs. To further verify our projection-aware RoPE, we replace the raw image indices RoPE with projection-aware RoPE in PRoPE~\cite{li2025prope}. As shown in Table~\ref{tab:ablation}, this modification improves overall performance, demonstrating that projection-aware RoPE is not specific to our overall architecture but can be applied to other camera-conditioning methods for heterogeneous-camera NVS.

%% file: table/main.tex
\begin{table}[t] \small
    \centering
    \caption{\textbf{NVS comparisons on ScanNet++.} In-distribution data evaluation. Our methods achieve best results across three protocols. }
    \label{tab:main}
    \setlength{\tabcolsep}{4pt}
    \resizebox{\linewidth}{!}{%
        \begin{tabular}{lccc ccc ccc ccc}
        \toprule
        \multirow{2}{*}{Methods} & \multirow{2}{*}{\shortstack{Rel.\\Camera\\PE}} & \multirow{2}{*}{Raymaps} & \multirow{2}{*}{\shortstack{2D\\RoPE}}
        & \multicolumn{3}{c}{Undistorted} & \multicolumn{3}{c}{Fisheye} & \multicolumn{3}{c}{Mixed} \\
        \cmidrule(lr){5-7} \cmidrule(lr){8-10} \cmidrule(lr){11-13}
        & & & & {PSNR $\uparrow$} & {SSIM $\uparrow$} & {LPIPS $\downarrow$} & {PSNR $\uparrow$} & {SSIM $\uparrow$} & {LPIPS $\downarrow$} & {PSNR $\uparrow$} & {SSIM $\uparrow$} & {LPIPS $\downarrow$} \\
        \midrule
        LVSM~\cite{jin2025lvsm} & - & Pl\"{u}cker & Raw & 24.716 & 0.781 & 0.174 & 24.290 & 0.763 & 0.175 & 24.104 & 0.766 & 0.183 \\
        PRoPE~\cite{li2025prope}& PRoPE & CamRay & Raw & 23.510 & 0.760 & 0.173 & 22.780 & 0.733 & 0.181 & 21.614 & 0.712 & 0.207 \\
        \midrule
        UCPE-N & Rel. Ray & - & Raw & 26.175 & 0.812 & 0.133 & 25.699 & 0.799 & 0.131 & 25.149 & 0.795 & 0.142 \\
        UCPE-P & Rel. Ray & Pl\"{u}cker & Raw & 26.083 & 0.811 & 0.136 & 25.568 & 0.797 & 0.135 & 25.204 & 0.796 & 0.144 \\
        UCPE-C & Rel. Ray & CamRay & Raw & 26.141 & 0.813 & 0.132 & 25.591 & 0.799 & 0.131 & 25.278 & 0.798 & 0.140 \\
        \midrule
        Ours (Raw) & Rel. Ray & Local Ray & Raw & \cellcolor{second}26.308 & \cellcolor{best}0.815 & \cellcolor{second}0.131 & \cellcolor{second}25.784 & \cellcolor{best}0.800 & \cellcolor{second}0.130 & \cellcolor{second}25.347 & \cellcolor{second}0.799 & \cellcolor{second}0.139 \\
        Ours (Full) & Rel. Ray & Local Ray & Proj. & \cellcolor{best}26.429 & \cellcolor{best}0.815 & \cellcolor{best}0.129 & \cellcolor{best}25.911 & \cellcolor{best}0.800 & \cellcolor{best}0.128 & \cellcolor{best}25.670 & \cellcolor{best}0.801 & \cellcolor{best}0.136 \\
        \bottomrule
        \end{tabular}
    }
    \vspace{-7pt}
\end{table}

%% file: table/pano.tex
\begin{table*}[t]
    \centering
    \begin{minipage}[t]{0.48\linewidth}
    \centering
    \small
    \captionof{table}{\textbf{Zero-shot NVS comparisons.} Evaluated on panoramic targets. Raymaps: N/P/C/L=None/Pl\"{u}cker/CamRay/Local Ray.}
    \label{tab:pano_zeroshot}
    \resizebox{\linewidth}{!}{%
    \begin{tabular}{cc ccc ccc}
    \toprule
    \multirow{2}{*}{Ray.} & \multirow{2}{*}{\shortstack{2D\\RoPE}}
    & \multicolumn{3}{c}{Virtual Cams $\rightarrow$ Panorama} & \multicolumn{3}{c}{DSLR $\rightarrow$ Panorama}\\
    \cmidrule(lr){3-5} \cmidrule(lr){6-8}
    & & {PSNR $\uparrow$} & {SSIM $\uparrow$} & {LPIPS $\downarrow$} & {PSNR $\uparrow$} & {SSIM $\uparrow$} & {LPIPS $\downarrow$} \\
    \midrule
    N & Raw & 22.426 & 0.731 & 0.167 & 22.659 & 0.765 & 0.237 \\
    P & Raw & 22.853 & 0.740 & 0.170 & 22.760 & 0.762 & 0.257\\
    C & Raw & Fail & Fail & Fail & Fail & Fail & Fail \\
    \midrule
    L & Raw & \cellcolor{second}23.139 & \cellcolor{second}0.755 & \cellcolor{best}0.141 & \cellcolor{best}22.999 & \cellcolor{best}0.774 & \cellcolor{best}0.226 \\
    L & Proj. & \cellcolor{best}23.346 & \cellcolor{best}0.760 & \cellcolor{second}0.142 & \cellcolor{second}22.815 & \cellcolor{second}0.769 & \cellcolor{second}0.232 \\
    \bottomrule
    \end{tabular}
    }
    \end{minipage}
    \hfill
    \begin{minipage}[t]{0.48\linewidth}
    \centering
    \small

        \captionof{table}{\textbf{Ablation studies.} Applying the projection-aware 2D RoPE to PRoPE-CamRay, where C/L=CamRay/Local Ray.}
        \label{tab:ablation}
    \resizebox{\linewidth}{!}{%
        \begin{tabular}{l lll lll}
        \toprule
        \multirow{2}{*}{2D RoPE} &
        \multicolumn{3}{c}{Fisheye} &
        \multicolumn{3}{c}{Mixed} \\
        \cmidrule(lr){2-4} \cmidrule(lr){5-7}
        & PSNR $\uparrow$ & SSIM $\uparrow$ & LPIPS $\downarrow$
        & PSNR $\uparrow$ & SSIM $\uparrow$ & LPIPS $\downarrow$ \\
        \midrule
        PRoPE(C)-Raw   & 22.780 & 0.733 & 0.181 & 21.614 & 0.712 & 0.207 \\
        PRoPE(C)-Proj. & 25.049 \textcolor{red}{+} & 0.781 \textcolor{red}{+} & 0.145 \textcolor{red}{-} & 24.856 \textcolor{red}{+} & 0.783 \textcolor{red}{+} & 0.153 \textcolor{red}{-} \\
        \midrule
        UCPE(L)-Raw & 25.784 & 0.800 & 0.130 & 25.347 & 0.799 & 0.139 \\
        UCPE(L)-Proj. & 25.911 \textcolor{red}{+} & 0.800 & 0.128 \textcolor{red}{-} & 25.670 \textcolor{red}{+} & 0.801 \textcolor{red}{+} & 0.136 \textcolor{red}{-} \\
        \bottomrule
        \end{tabular}
    }

    \end{minipage}
    \vspace{-10pt}
\end{table*}

%% file: pics/main.tex
\begin{figure}[t]
    \centering
    \includegraphics[width=1.0\textwidth]{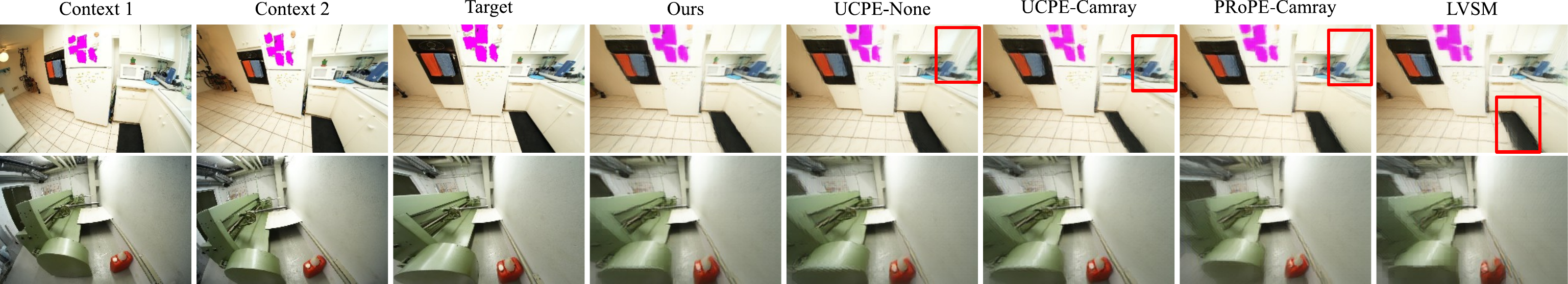}
    \caption{\textbf{Qualitative comparisons on mixed camera evaluation.}
    Our method and absolute raymap baselines (LVSM) have capabilities to infer a correct field of view across different camera types.  Compared with ours, all other methods produce blurred results under heterogeneous-camera settings. }
    \label{fig:main}
    \vspace{-5pt}
\end{figure}

%% file: pics/pano.tex
\begin{figure}[t]
    \centering
    \includegraphics[width=0.8\textwidth]{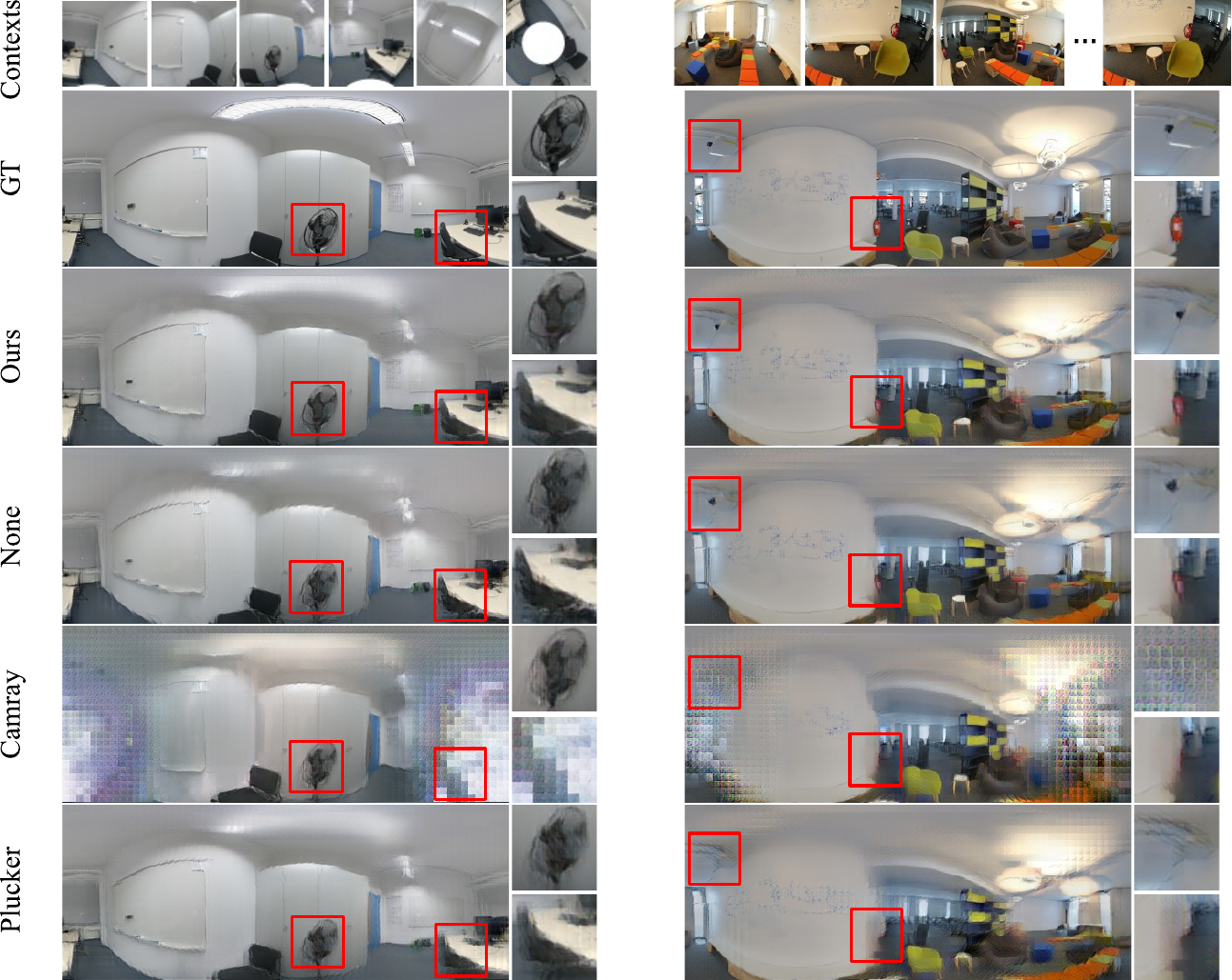}
    \caption{\textbf{Qualitative comparisons on zero-shot panoramic evaluation.}
    Left: virtual cameras to panorama images; Right: DSLR to panorama. Our methods exhibit less artifacts in the zero-shot generalization to panoramic targets.}
    \label{fig:pano}
    \vspace{-10pt}
\end{figure}

%% file: section/5_conclusion.tex
\section{Conclusion}\label{sec:conclusion}

We studied feed-forward NVS in a heterogeneous-camera setting, where different projection families are handled by a single model rather than separate camera-specific systems. Since camera variation is continuous and difficult to cover exhaustively with calibrated heterogeneous-camera data, we argue that generalization should come from reusable geometric units rather than memorized camera types. By exposing token-local ray distributions and ray-domain positional coordinates, our method improves mixed-camera synthesis and exhibits zero-shot generalization to unseen target cameras.

\paragraph{Limitations.}
Heterogeneous-camera NVS also reflects a broader data challenge in the field. Our method relies on calibrated datasets, yet large-scale captures with diverse, accurately calibrated camera systems are difficult to collect and curate. While our results show that reusable geometric representations can reduce reliance on camera-specific supervision, richer heterogeneous-camera data would further improve robustness and help scale this setting to multi-sensor applications.

%% file: section/x_supp.tex
\section{Implementation Details}
\subsection{Architecture}
We use the lightweight decoder-only LVSM backbone~\cite{jin2025lvsm} for all
methods. Each model has 6 Transformer layers, 16 attention heads, a token
dimension of 768, and patch size 8, resulting in approximately 24.5M
parameters. Context views and the target camera are converted into patch tokens
and concatenated into one sequence before self-attention.

\paragraph{Optimization.}
We follow the PRoPE training recipe~\cite{li2025prope}: all methods are trained
for 80k iterations with a total batch size of 8 on two NVIDIA A100 GPUs. We use
a cosine learning-rate scheduler with 2.5k warmup iterations. Unless otherwise
specified in the experiment section, each training sample contains two context
views and one target view, and all baselines use the same backbone, training
schedule, image resolution, and view sampling protocol.



\subsection{DSLR-to-Panorama sampling and evaluation pair construction}
\label{sec:dslr_pano_pairs}

We evaluate zero-shot panorama prediction by constructing
``DSLR$\rightarrow$panorama'' pairs in ScanNet++. For each scene, every panocam
frame is treated as a target panorama $p$, and we select a fixed-size set of
$K$ DSLR context images,
\begin{equation}
\mathcal{C}(p)=\{i_1,\ldots,i_K\}.
\end{equation}
Our sampling objective is twofold: (1) context views should be spatially close to
the target panorama, and (2) their viewing directions should cover as much of the
target panorama's spherical field of view as possible.

\paragraph{Overall procedure.}
For each target panorama $p$, we construct $\mathcal{C}(p)$ with the following
geometry-guided pipeline. First, we keep only DSLR frames whose camera centers
are within a distance threshold of the panorama center. Second, we transform each
remaining DSLR viewing direction into the local coordinate frame of the target
panorama. Third, we discretize the panorama sphere into angular bins and compute
which bins each DSLR image covers. Finally, we greedily add the DSLR image that
contributes the largest amount of previously uncovered solid angle. The greedy
selection stops once $K$ context images have been selected. During evaluation, we
only keep target panoramas whose final coverage exceeds a prescribed threshold,
so every retained pair has exactly $K$ inputs and sufficient angular support.

\paragraph{Candidate filtering.}
Let $T_p=[R_p|t_p]$ denote the pose of target panorama $p$, where $R_p$ is the
camera-to-world rotation and $t_p$ is the camera center. Let
$T_i=[R_i|t_i]$ denote the pose of a DSLR frame $i$ from the same scene. We first
filter DSLR candidates by Euclidean distance between camera centers:
\begin{equation}
\mathcal{S}(p)=\Big\{i:\ \|t_i-t_p\|\le d_{\max}\Big\}.
\label{eq:supp_candidate_set}
\end{equation}

\paragraph{Viewing direction in the panorama frame.}
For each candidate $i\in\mathcal{S}(p)$, we use the DSLR camera forward axis as
its viewing direction. With $e_z=[0,0,1]^\top$, the forward direction in world
coordinates is
\begin{equation}
d_i^{\mathrm{world}}=R_i e_z.
\label{eq:supp_world_direction}
\end{equation}
We then express this direction in the local coordinate frame of the target
panorama:
\begin{equation}
d_i^{p}=R_p^\top d_i^{\mathrm{world}}.
\label{eq:supp_pano_direction}
\end{equation}
Writing $d_i^{p}=(x_i,y_i,z_i)^\top$, we convert it to azimuth and elevation as
\begin{equation}
\mathrm{az}_i=\operatorname{atan2}(x_i,z_i),\qquad
\mathrm{el}_i=\arcsin\!\left(\frac{y_i}{\|d_i^{p}\|_2}\right).
\label{eq:supp_az_el}
\end{equation}

\paragraph{Angular coverage.}
We discretize the panorama domain into angular bins
$\mathcal{B}=\{b\}$. Each bin $b$ has a center
$(\mathrm{az}_b,\mathrm{el}_b)$ and a non-negative weight $w_b$. We define a
binary coverage indicator
\begin{equation}
\mathrm{cov}(i,b)=
\begin{cases}
1, & \text{if DSLR view } i \text{ covers angular bin } b,\\
0, & \text{otherwise}.
\end{cases}
\label{eq:supp_cov_indicator}
\end{equation}
In practice, $\mathrm{cov}(i,b)=1$ when the bin center lies inside the DSLR
camera field of view after projection from the panorama frame to the DSLR frame.
Given a partial context set $\mathcal{C}$, the set of angular bins already
covered by $\mathcal{C}$ is
\begin{equation}
\mathrm{Covered}(\mathcal{C})
=\Big\{b\in\mathcal{B}:\ \max_{i\in\mathcal{C}}\mathrm{cov}(i,b)=1\Big\}.
\label{eq:supp_covered_set}
\end{equation}
The weighted coverage fraction is
\begin{equation}
\mathrm{CovFrac}(\mathcal{C})
=
\frac{\sum_{b\in\mathrm{Covered}(\mathcal{C})} w_b}
{\sum_{b\in\mathcal{B}} w_b}.
\label{eq:supp_coverage_fraction}
\end{equation}

\paragraph{Greedy context selection.}
Starting from $\mathcal{C}_0=\emptyset$, we add one DSLR image at a time. At
step $m$, the next image is selected by maximizing the newly covered weighted
area:
\begin{equation}
i_m^\star
=
\arg\max_{i\in \mathcal{S}(p)\setminus \mathcal{C}_{m-1}}
\Delta(i\mid\mathcal{C}_{m-1}),
\label{eq:supp_greedy_selection}
\end{equation}
where the marginal gain is
\begin{equation}
\Delta(i\mid\mathcal{C})
=
\sum_{b\in\mathcal{B}} w_b\,
\mathbb{I}\!\left[
\mathrm{cov}(i,b)=1
\ \wedge\
b\notin\mathrm{Covered}(\mathcal{C})
\right].
\label{eq:supp_marginal_gain}
\end{equation}
We then update
\begin{equation}
\mathcal{C}_m=\mathcal{C}_{m-1}\cup\{i_m^\star\},
\qquad m=1,\ldots,K.
\label{eq:supp_context_update}
\end{equation}
The resulting set $\mathcal{C}(p)=\mathcal{C}_K$ is used as the context input
for target panorama $p$ if it passes the final coverage threshold.

\paragraph{Solid-angle weighting.}
We use solid-angle weights so that uniformly spaced bins in
$(\mathrm{az},\mathrm{el})$ do not over-emphasize regions near the poles. For a bin $b$ centered at elevation $\mathrm{el}_b$, we therefore set
\begin{equation}
w_b=\max\!\big(0,\cos(\mathrm{el}_b)\big).
\label{eq:supp_bin_weight}
\end{equation}
Since $\mathrm{CovFrac}$ in Eq.~\eqref{eq:supp_coverage_fraction} normalizes by
the total weight, the absolute scale of $w_b$ does not affect the final coverage
ratio.

\paragraph{Default hyper-parameters.}
The precomputed evaluation manifests use $5^\circ$ bins in both azimuth and
elevation. We set the panorama--DSLR center-distance threshold to
$d_{\max}=2.0\,\mathrm{m}$, select $K=10$ context DSLR images
per target panorama, and retain only targets whose final coverage satisfies
$\mathrm{CovFrac}(\mathcal{C}(p))>0.8$. This filtering results in 36 valid scenes in total.

\section{More qualitative results}
\input{pics/main_supp.tex}

\section{Assets License} \label{license}
In Table~\ref{tbl:license}, we list the licenses of all the existing assets including the code and data we have used in this work.

\begin{table*}[ht] \setlength{\tabcolsep}{1mm}
  \centering
  \scriptsize
  \begin{tabular}{lc}
    \toprule
     Asset & License Link\\
    \midrule
    LVSM~\cite{jin2025lvsm} & \href{https://github.com/Haian-Jin/LVSM/blob/main/LICENSE.md}{https://github.com/Haian-Jin/LVSM/blob/main/LICENSE.md} \\
    ScanNet++ v2~\cite{yeshwanthliu2023scannetpp} & \href{https://scannetpp.mlsg.cit.tum.de/scannetpp/static/scannetpp-terms-of-use.pdf}{https://scannetpp.mlsg.cit.tum.de/scannetpp/static/scannetpp-terms-of-use.pdf} \\
    UCPE~\cite{zhang2025ucpe} & \href{https://github.com/chengzhag/UCPE?tab=MIT-1-ov-file}{https://github.com/chengzhag/UCPE?tab=MIT-1-ov-file} \\
    \bottomrule
  \end{tabular}
  \vspace{2mm}
  \caption{\textbf{Licenses of existing assets we have used in this work.}}
  \label{tbl:license}
\end{table*}

%% file: pics/main_supp.tex
\begin{figure}[t]
    \centering
    \includegraphics[width=1.0\textwidth]{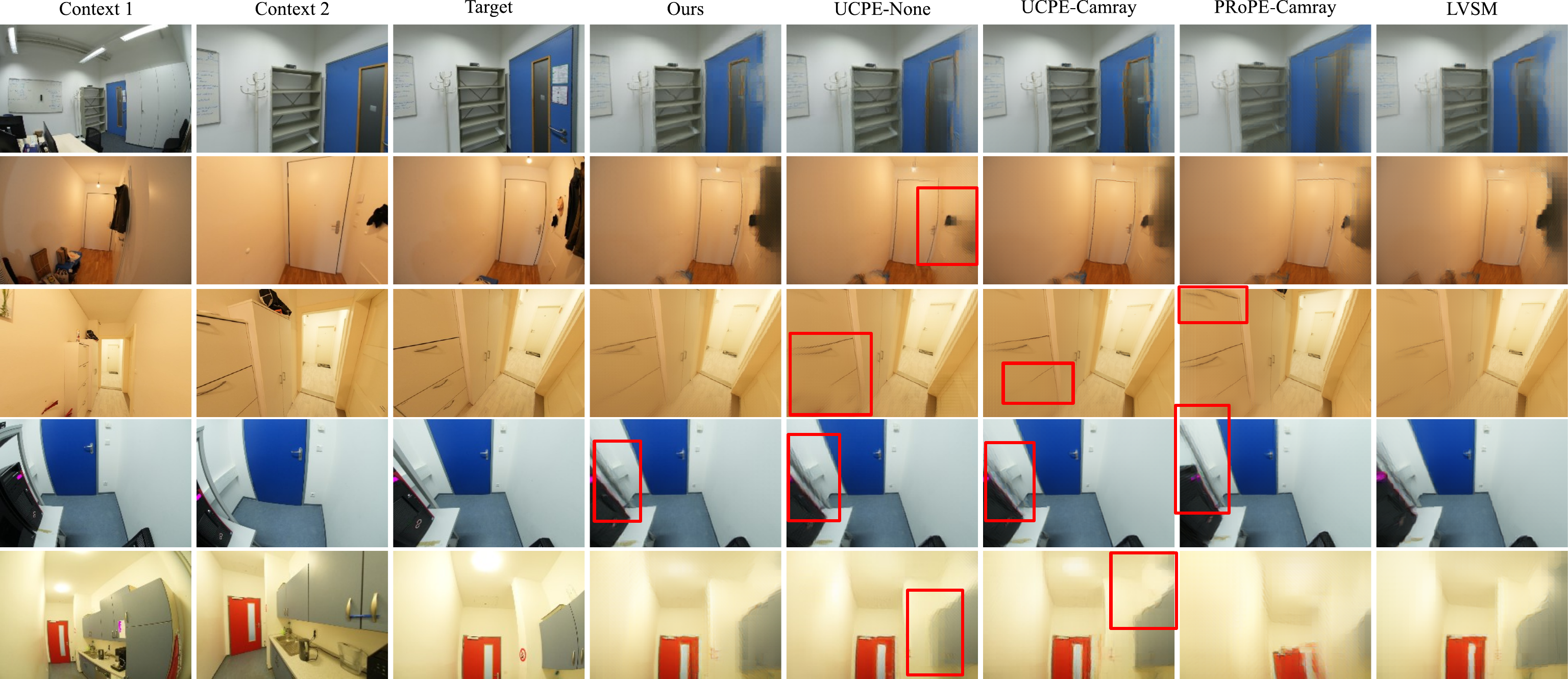}
    \caption{\textbf{More qualitative comparisons on mixed camera evaluation.} }
    \label{fig:supp_main}
    \vspace{-5pt}
\end{figure}

%% file: checklist.tex
\section*{NeurIPS Paper Checklist}

\begin{enumerate}

\item {\bf Claims}
    \item[] Question: Do the main claims made in the abstract and introduction accurately reflect the paper's contributions and scope?
    \item[] Answer: \answerYes{}
    \item[] Justification: The abstract and introduction state the heterogeneous-camera NVS setting, the local raymap and projection-aware RoPE contributions, and the claimed empirical gains. These claims are supported by the method and experiments in Sections~\ref{sec:method} and~\ref{sec:experiments}.
    \item[] Guidelines:
    \begin{itemize}
        \item The answer \answerNA{} means that the abstract and introduction do not include the claims made in the paper.
        \item The abstract and/or introduction should clearly state the claims made, including the contributions made in the paper and important assumptions and limitations. A \answerNo{} or \answerNA{} answer to this question will not be perceived well by the reviewers. 
        \item The claims made should match theoretical and experimental results, and reflect how much the results can be expected to generalize to other settings. 
        \item It is fine to include aspirational goals as motivation as long as it is clear that these goals are not attained by the paper. 
    \end{itemize}

\item {\bf Limitations}
    \item[] Question: Does the paper discuss the limitations of the work performed by the authors?
    \item[] Answer: \answerYes{}
    \item[] Justification: The paper discusses limitations in the conclusion.
    \item[] Guidelines:
    \begin{itemize}
        \item The answer \answerNA{} means that the paper has no limitation while the answer \answerNo{} means that the paper has limitations, but those are not discussed in the paper. 
        \item The authors are encouraged to create a separate ``Limitations'' section in their paper.
        \item The paper should point out any strong assumptions and how robust the results are to violations of these assumptions (e.g., independence assumptions, noiseless settings, model well-specification, asymptotic approximations only holding locally). The authors should reflect on how these assumptions might be violated in practice and what the implications would be.
        \item The authors should reflect on the scope of the claims made, e.g., if the approach was only tested on a few datasets or with a few runs. In general, empirical results often depend on implicit assumptions, which should be articulated.
        \item The authors should reflect on the factors that influence the performance of the approach. For example, a facial recognition algorithm may perform poorly when image resolution is low or images are taken in low lighting. Or a speech-to-text system might not be used reliably to provide closed captions for online lectures because it fails to handle technical jargon.
        \item The authors should discuss the computational efficiency of the proposed algorithms and how they scale with dataset size.
        \item If applicable, the authors should discuss possible limitations of their approach to address problems of privacy and fairness.
        \item While the authors might fear that complete honesty about limitations might be used by reviewers as grounds for rejection, a worse outcome might be that reviewers discover limitations that aren't acknowledged in the paper. The authors should use their best judgment and recognize that individual actions in favor of transparency play an important role in developing norms that preserve the integrity of the community. Reviewers will be specifically instructed to not penalize honesty concerning limitations.
    \end{itemize}

\item {\bf Theory assumptions and proofs}
    \item[] Question: For each theoretical result, does the paper provide the full set of assumptions and a complete (and correct) proof?
    \item[] Answer: \answerNA{}
    \item[] Justification: The paper does not present theoretical results, but present emperical results.
    \item[] Guidelines:
    \begin{itemize}
        \item The answer \answerNA{} means that the paper does not include theoretical results. 
        \item All the theorems, formulas, and proofs in the paper should be numbered and cross-referenced.
        \item All assumptions should be clearly stated or referenced in the statement of any theorems.
        \item The proofs can either appear in the main paper or the supplemental material, but if they appear in the supplemental material, the authors are encouraged to provide a short proof sketch to provide intuition. 
        \item Inversely, any informal proof provided in the core of the paper should be complemented by formal proofs provided in appendix or supplemental material.
        \item Theorems and Lemmas that the proof relies upon should be properly referenced. 
    \end{itemize}

    \item {\bf Experimental result reproducibility}
    \item[] Question: Does the paper fully disclose all the information needed to reproduce the main experimental results of the paper to the extent that it affects the main claims and/or conclusions of the paper (regardless of whether the code and data are provided or not)?
    \item[] Answer: \answerYes{}
    \item[] Justification: Section~\ref{sec:experiments} specifies the dataset and training protocals used in the experiments. More details about architectures and evaluation protocals are included in the supplementary.
    \item[] Guidelines:
    \begin{itemize}
        \item The answer \answerNA{} means that the paper does not include experiments.
        \item If the paper includes experiments, a \answerNo{} answer to this question will not be perceived well by the reviewers: Making the paper reproducible is important, regardless of whether the code and data are provided or not.
        \item If the contribution is a dataset and\slash or model, the authors should describe the steps taken to make their results reproducible or verifiable. 
        \item Depending on the contribution, reproducibility can be accomplished in various ways. For example, if the contribution is a novel architecture, describing the architecture fully might suffice, or if the contribution is a specific model and empirical evaluation, it may be necessary to either make it possible for others to replicate the model with the same dataset, or provide access to the model. In general. releasing code and data is often one good way to accomplish this, but reproducibility can also be provided via detailed instructions for how to replicate the results, access to a hosted model (e.g., in the case of a large language model), releasing of a model checkpoint, or other means that are appropriate to the research performed.
        \item While NeurIPS does not require releasing code, the conference does require all submissions to provide some reasonable avenue for reproducibility, which may depend on the nature of the contribution. For example
        \begin{enumerate}
            \item If the contribution is primarily a new algorithm, the paper should make it clear how to reproduce that algorithm.
            \item If the contribution is primarily a new model architecture, the paper should describe the architecture clearly and fully.
            \item If the contribution is a new model (e.g., a large language model), then there should either be a way to access this model for reproducing the results or a way to reproduce the model (e.g., with an open-source dataset or instructions for how to construct the dataset).
            \item We recognize that reproducibility may be tricky in some cases, in which case authors are welcome to describe the particular way they provide for reproducibility. In the case of closed-source models, it may be that access to the model is limited in some way (e.g., to registered users), but it should be possible for other researchers to have some path to reproducing or verifying the results.
        \end{enumerate}
    \end{itemize}

\item {\bf Open access to data and code}
    \item[] Question: Does the paper provide open access to the data and code, with sufficient instructions to faithfully reproduce the main experimental results, as described in supplemental material?
    \item[] Answer: \answerNo{}
    \item[] Justification: The submission does not currently provide public codes. These will be released after publication. We use a publicly avaiable dataset.
    \item[] Guidelines:
    \begin{itemize}
        \item The answer \answerNA{} means that paper does not include experiments requiring code.
        \item Please see the NeurIPS code and data submission guidelines (\url{https://neurips.cc/public/guides/CodeSubmissionPolicy}) for more details.
        \item While we encourage the release of code and data, we understand that this might not be possible, so \answerNo{} is an acceptable answer. Papers cannot be rejected simply for not including code, unless this is central to the contribution (e.g., for a new open-source benchmark).
        \item The instructions should contain the exact command and environment needed to run to reproduce the results. See the NeurIPS code and data submission guidelines (\url{https://neurips.cc/public/guides/CodeSubmissionPolicy}) for more details.
        \item The authors should provide instructions on data access and preparation, including how to access the raw data, preprocessed data, intermediate data, and generated data, etc.
        \item The authors should provide scripts to reproduce all experimental results for the new proposed method and baselines. If only a subset of experiments are reproducible, they should state which ones are omitted from the script and why.
        \item At submission time, to preserve anonymity, the authors should release anonymized versions (if applicable).
        \item Providing as much information as possible in supplemental material (appended to the paper) is recommended, but including URLs to data and code is permitted.
    \end{itemize}

\item {\bf Experimental setting/details}
    \item[] Question: Does the paper specify all the training and test details (e.g., data splits, hyperparameters, how they were chosen, type of optimizer) necessary to understand the results?
    \item[] Answer: \answerYes{}
    \item[] Justification: The experimental setup and supplementary describe these details.
    \item[] Guidelines:
    \begin{itemize}
        \item The answer \answerNA{} means that the paper does not include experiments.
        \item The experimental setting should be presented in the core of the paper to a level of detail that is necessary to appreciate the results and make sense of them.
        \item The full details can be provided either with the code, in appendix, or as supplemental material.
    \end{itemize}

\item {\bf Experiment statistical significance}
    \item[] Question: Does the paper report error bars suitably and correctly defined or other appropriate information about the statistical significance of the experiments?
    \item[] Answer: \answerNo{}
    \item[] Justification: The current experiments report metrics without error bars or statistical significance tests. The experiments are evaluated at sufficient samples of data.
    \item[] Guidelines:
    \begin{itemize}
        \item The answer \answerNA{} means that the paper does not include experiments.
        \item The authors should answer \answerYes{} if the results are accompanied by error bars, confidence intervals, or statistical significance tests, at least for the experiments that support the main claims of the paper.
        \item The factors of variability that the error bars are capturing should be clearly stated (for example, train/test split, initialization, random drawing of some parameter, or overall run with given experimental conditions).
        \item The method for calculating the error bars should be explained (closed form formula, call to a library function, bootstrap, etc.)
        \item The assumptions made should be given (e.g., Normally distributed errors).
        \item It should be clear whether the error bar is the standard deviation or the standard error of the mean.
        \item It is OK to report 1-sigma error bars, but one should state it. The authors should preferably report a 2-sigma error bar than state that they have a 96\% CI, if the hypothesis of Normality of errors is not verified.
        \item For asymmetric distributions, the authors should be careful not to show in tables or figures symmetric error bars that would yield results that are out of range (e.g., negative error rates).
        \item If error bars are reported in tables or plots, the authors should explain in the text how they were calculated and reference the corresponding figures or tables in the text.
    \end{itemize}

\item {\bf Experiments compute resources}
    \item[] Question: For each experiment, does the paper provide sufficient information on the computer resources (type of compute workers, memory, time of execution) needed to reproduce the experiments?
    \item[] Answer: \answerYes{}
    \item[] Justification: The paper reports the compute setup, including two NVIDIA A100 GPUs, total batch size, and training iterations. Our methods aligned with academic resources and are not constrained by the CPU or memory.
    \item[] Guidelines:
    \begin{itemize}
        \item The answer \answerNA{} means that the paper does not include experiments.
        \item The paper should indicate the type of compute workers CPU or GPU, internal cluster, or cloud provider, including relevant memory and storage.
        \item The paper should provide the amount of compute required for each of the individual experimental runs as well as estimate the total compute. 
        \item The paper should disclose whether the full research project required more compute than the experiments reported in the paper (e.g., preliminary or failed experiments that didn't make it into the paper). 
    \end{itemize}
    
\item {\bf Code of ethics}
    \item[] Question: Does the research conducted in the paper conform, in every respect, with the NeurIPS Code of Ethics \url{https://neurips.cc/public/EthicsGuidelines}?
    \item[] Answer: \answerYes{}
    \item[] Justification: The research uses existing research datasets and evaluates view synthesis methods without human-subject experiments or unsafe data release. We are not aware of any deviation from the NeurIPS Code of Ethics.
    \item[] Guidelines:
    \begin{itemize}
        \item The answer \answerNA{} means that the authors have not reviewed the NeurIPS Code of Ethics.
        \item If the authors answer \answerNo, they should explain the special circumstances that require a deviation from the Code of Ethics.
        \item The authors should make sure to preserve anonymity (e.g., if there is a special consideration due to laws or regulations in their jurisdiction).
    \end{itemize}

\item {\bf Broader impacts}
    \item[] Question: Does the paper discuss both potential positive societal impacts and negative societal impacts of the work performed?
    \item[] Answer: \answerNA{}
    \item[] Justification: There is no societal impact of the work performed.
    \item[] Guidelines:
    \begin{itemize}
        \item The answer \answerNA{} means that there is no societal impact of the work performed.
        \item If the authors answer \answerNA{} or \answerNo, they should explain why their work has no societal impact or why the paper does not address societal impact.
        \item Examples of negative societal impacts include potential malicious or unintended uses (e.g., disinformation, generating fake profiles, surveillance), fairness considerations (e.g., deployment of technologies that could make decisions that unfairly impact specific groups), privacy considerations, and security considerations.
        \item The conference expects that many papers will be foundational research and not tied to particular applications, let alone deployments. However, if there is a direct path to any negative applications, the authors should point it out. For example, it is legitimate to point out that an improvement in the quality of generative models could be used to generate Deepfakes for disinformation. On the other hand, it is not needed to point out that a generic algorithm for optimizing neural networks could enable people to train models that generate Deepfakes faster.
        \item The authors should consider possible harms that could arise when the technology is being used as intended and functioning correctly, harms that could arise when the technology is being used as intended but gives incorrect results, and harms following from (intentional or unintentional) misuse of the technology.
        \item If there are negative societal impacts, the authors could also discuss possible mitigation strategies (e.g., gated release of models, providing defenses in addition to attacks, mechanisms for monitoring misuse, mechanisms to monitor how a system learns from feedback over time, improving the efficiency and accessibility of ML).
    \end{itemize}
    
\item {\bf Safeguards}
    \item[] Question: Does the paper describe safeguards that have been put in place for responsible release of data or models that have a high risk for misuse (e.g., pre-trained language models, image generators, or scraped datasets)?
    \item[] Answer: \answerNA{}
    \item[] Justification: The paper does not release a high-risk pretrained generative model or scraped dataset. The work evaluates a view-synthesis method on existing research data.
    \item[] Guidelines:
    \begin{itemize}
        \item The answer \answerNA{} means that the paper poses no such risks.
        \item Released models that have a high risk for misuse or dual-use should be released with necessary safeguards to allow for controlled use of the model, for example by requiring that users adhere to usage guidelines or restrictions to access the model or implementing safety filters. 
        \item Datasets that have been scraped from the Internet could pose safety risks. The authors should describe how they avoided releasing unsafe images.
        \item We recognize that providing effective safeguards is challenging, and many papers do not require this, but we encourage authors to take this into account and make a best faith effort.
    \end{itemize}

\item {\bf Licenses for existing assets}
    \item[] Question: Are the creators or original owners of assets (e.g., code, data, models), used in the paper, properly credited and are the license and terms of use explicitly mentioned and properly respected?
    \item[] Answer: \answerYes{}
    \item[] Justification: The paper cites the datasets and prior methods used in the experiments and list license in the supplementary.
    \item[] Guidelines:
    \begin{itemize}
        \item The answer \answerNA{} means that the paper does not use existing assets.
        \item The authors should cite the original paper that produced the code package or dataset.
        \item The authors should state which version of the asset is used and, if possible, include a URL.
        \item The name of the license (e.g., CC-BY 4.0) should be included for each asset.
        \item For scraped data from a particular source (e.g., website), the copyright and terms of service of that source should be provided.
        \item If assets are released, the license, copyright information, and terms of use in the package should be provided. For popular datasets, \url{paperswithcode.com/datasets} has curated licenses for some datasets. Their licensing guide can help determine the license of a dataset.
        \item For existing datasets that are re-packaged, both the original license and the license of the derived asset (if it has changed) should be provided.
        \item If this information is not available online, the authors are encouraged to reach out to the asset's creators.
    \end{itemize}

\item {\bf New assets}
    \item[] Question: Are new assets introduced in the paper well documented and is the documentation provided alongside the assets?
    \item[] Answer: \answerNA{}
    \item[] Justification: The paper does not introduce or release a new dataset.
    \item[] Guidelines:
    \begin{itemize}
        \item The answer \answerNA{} means that the paper does not release new assets.
        \item Researchers should communicate the details of the dataset\slash code\slash model as part of their submissions via structured templates. This includes details about training, license, limitations, etc. 
        \item The paper should discuss whether and how consent was obtained from people whose asset is used.
        \item At submission time, remember to anonymize your assets (if applicable). You can either create an anonymized URL or include an anonymized zip file.
    \end{itemize}

\item {\bf Crowdsourcing and research with human subjects}
    \item[] Question: For crowdsourcing experiments and research with human subjects, does the paper include the full text of instructions given to participants and screenshots, if applicable, as well as details about compensation (if any)? 
    \item[] Answer: \answerNA{}
    \item[] Justification: The work does not involve crowdsourcing experiments or research with human subjects.
    \item[] Guidelines:
    \begin{itemize}
        \item The answer \answerNA{} means that the paper does not involve crowdsourcing nor research with human subjects.
        \item Including this information in the supplemental material is fine, but if the main contribution of the paper involves human subjects, then as much detail as possible should be included in the main paper. 
        \item According to the NeurIPS Code of Ethics, workers involved in data collection, curation, or other labor should be paid at least the minimum wage in the country of the data collector. 
    \end{itemize}

\item {\bf Institutional review board (IRB) approvals or equivalent for research with human subjects}
    \item[] Question: Does the paper describe potential risks incurred by study participants, whether such risks were disclosed to the subjects, and whether Institutional Review Board (IRB) approvals (or an equivalent approval/review based on the requirements of your country or institution) were obtained?
    \item[] Answer: \answerNA{}
    \item[] Justification: The work does not involve crowdsourcing experiments or research with human subjects, so IRB approval or equivalent review is not applicable.
    \item[] Guidelines:
    \begin{itemize}
        \item The answer \answerNA{} means that the paper does not involve crowdsourcing nor research with human subjects.
        \item Depending on the country in which research is conducted, IRB approval (or equivalent) may be required for any human subjects research. If you obtained IRB approval, you should clearly state this in the paper. 
        \item We recognize that the procedures for this may vary significantly between institutions and locations, and we expect authors to adhere to the NeurIPS Code of Ethics and the guidelines for their institution. 
        \item For initial submissions, do not include any information that would break anonymity (if applicable), such as the institution conducting the review.
    \end{itemize}

\item {\bf Declaration of LLM usage}
    \item[] Question: Does the paper describe the usage of LLMs if it is an important, original, or non-standard component of the core methods in this research? Note that if the LLM is used only for writing, editing, or formatting purposes and does \emph{not} impact the core methodology, scientific rigor, or originality of the research, declaration is not required.
    \item[] Answer: \answerNA{}
    \item[] Justification: LLMs are not an important, original, or non-standard component of the core method or experiments.
    \item[] Guidelines:
    \begin{itemize}
        \item The answer \answerNA{} means that the core method development in this research does not involve LLMs as any important, original, or non-standard components.
        \item Please refer to our LLM policy in the NeurIPS handbook for what should or should not be described.
    \end{itemize}

\end{enumerate}

%% file: foundation.bib
@article{li2025prope,
    title={Cameras as Relative Positional Encoding},
    author={Li, Ruilong and Yi, Brent and Liu, Junchen and Gao, Hang and Ma, Yi and Kanazawa, Angjoo},
    journal={Advances in Neural Information Processing Systems},
    year={2025}
}

@article{wu2026rayrope,
  title={RayRoPE: Projective Ray Positional Encoding for Multi-view Attention},
  author={Wu, Yu and Jeon, Minsik and Chang, Jen-Hao Rick and Tuzel, Oncel and Tulsiani, Shubham},
  journal={arXiv preprint arXiv:2601.15275},
  year={2026}
}

@article{zhang2025ucpe,
  title={Unified Camera Positional Encoding for Controlled Video Generation},
  author={Zhang, Cheng and Li, Boying and Wei, Meng and Cao, Yan-Pei and Gambardella, Camilo Cruz and Phung, Dinh and Cai, Jianfei},
  journal={arXiv preprint arXiv:2512.07237},
  year={2025}
}

@inproceedings{Miyato2024GTA,
    title={GTA: A Geometry-Aware Attention Mechanism for Multi-View Transformers},
    author={Miyato,Takeru and Jaeger, Bernhard and Welling, Max and Geiger, Andreas},
    booktitle={International Conference on Learning Representations (ICLR)},
    year={2024}
}

@inproceedings{Xiong2023CAPE,
  title={CAPE: Camera View Position Embedding for Multi-View 3D Object Detection},
  author={Xiong, Kaixin and Gong, Shi and Ye, Xiaoqing and Tan, Xiao and Wan, Ji and Ding, Errui and Wang, Jingdong and Bai, Xiang},
  booktitle = {Proceedings of the IEEE/CVF Conference on Computer Vision and Pattern Recognition},
  year={2023}
}

@article{su2024roformer,
  title={Roformer: Enhanced transformer with rotary position embedding},
  author={Su, Jianlin and Ahmed, Murtadha and Lu, Yu and Pan, Shengfeng and Bo, Wen and Liu, Yunfeng},
  journal={Neurocomputing},
  volume={568},
  pages={127063},
  year={2024},
  publisher={Elsevier}
}

@article{duan2026fisheye3r,
  title={Fisheye3R: Adapting Unified 3D Feed-Forward Foundation Models to Fisheye Lenses},
  author={Duan, Ruxiao and Hong, Erin and Zhao, Dongxu and Turner, Eric and Wong, Alex and Zhou, Yunwen},
  journal={arXiv preprint arXiv:2603.28896},
  year={2026}
}

@inproceedings{charatan2023pixelsplat,
  title     = {pixelsplat: 3d gaussian splats from image pairs for scalable generalizable 3d reconstruction},
  author    = {Charatan, David and Li, Sizhe and Tagliasacchi, Andrea and Sitzmann, Vincent},
  booktitle = {Proceedings of the IEEE/CVF Conference on Computer Vision and Pattern Recognition},
  year      = {2024}
}

@inproceedings{chen2024mvsplat,
  title   = {MVSplat: Efficient 3D Gaussian Splatting from Sparse Multi-View Images},
  author  = {Chen, Yuedong and Xu, Haofei and Zheng, Chuanxia and Zhuang, Bohan and Pollefeys, Marc and Geiger, Andreas and Cham, Tat-Jen and Cai, Jianfei},
  booktitle={European Conference on Computer Vision},
  year={2024},
  organization={Springer}
}

@inproceedings{chen2024mvsplat360,
    title     = {MVSplat360: Feed-Forward 360 Scene Synthesis from Sparse Views},
    author    = {Chen, Yuedong and Zheng, Chuanxia and Xu, Haofei and Zhuang, Bohan and Vedaldi, Andrea and Cham, Tat-Jen and Cai, Jianfei},
  booktitle={The Thirty-eighth Annual Conference on Neural Information Processing Systems},
    year      = {2024},
}

@inproceedings{zhang2024pansplat,
  title={PanSplat: 4K Panorama Synthesis with Feed-Forward Gaussian Splatting}, 
  author={Cheng Zhang and Haofei Xu and Qianyi Wu and Camilo Cruz Gambardella and Dinh Phung and Jianfei Cai},
  booktitle={Proceedings of the IEEE/CVF Conference on Computer Vision and Pattern Recognition},
  year={2025}
}

@inproceedings{jin2025lvsm,
  title={LVSM: A Large View Synthesis Model with Minimal 3D Inductive Bias},
  author={Haian Jin and Hanwen Jiang and Hao Tan and Kai Zhang and Sai Bi and Tianyuan Zhang and Fujun Luan and Noah Snavely and Zexiang Xu},
  booktitle={The Thirteenth International Conference on Learning Representations},
  year={2025}
}

@inproceedings{attention,
  author = {Vaswani, Ashish and Shazeer, Noam and Parmar, Niki and Uszkoreit, Jakob and Jones, Llion and Gomez, Aidan N. and Kaiser, \L{}ukasz and Polosukhin, Illia},
  title = {Attention is all you need},
  year = {2017},
  isbn = {9781510860964},
  publisher = {Curran Associates Inc.},
  address = {Red Hook, NY, USA},
  booktitle = {Proceedings of the 31st International Conference on Neural Information Processing Systems},
  pages = {6000–6010},
  numpages = {11},
  location = {Long Beach, California, USA},
  series = {NIPS'17}
}

@inproceedings{yu2021pixelnerf,
  title={pixelnerf: Neural radiance fields from one or few images},
  author={Yu, Alex and Ye, Vickie and Tancik, Matthew and Kanazawa, Angjoo},
  booktitle = {Proceedings of the IEEE/CVF Conference on Computer Vision and Pattern Recognition},
  pages={4578--4587},
  year={2021}
}

@inproceedings{chen2021mvsnerf,
  title={Mvsnerf: Fast generalizable radiance field reconstruction from multi-view stereo},
  author={Chen, Anpei and Xu, Zexiang and Zhao, Fuqiang and Zhang, Xiaoshuai and Xiang, Fanbo and Yu, Jingyi and Su, Hao},
  booktitle={Proceedings of the IEEE/CVF international conference on computer vision},
  pages={14124--14133},
  year={2021}
}

@article{smart2024splatt3r,
  title={Splatt3R: Zero-shot Gaussian Splatting from Uncalibrated Image Pairs}, 
  author={Brandon Smart and Chuanxia Zheng and Iro Laina and Victor Adrian Prisacariu},
  year={2024},
  eprint={2408.13912},
  archivePrefix={arXiv},
  primaryClass={cs.CV},
  journal={arXiv preprint arXiv:2408.13912},
}

@inproceedings{liu2021swinTrans,
  title={Swin transformer: Hierarchical vision transformer using shifted windows},
  author={Liu, Ze and Lin, Yutong and Cao, Yue and Hu, Han and Wei, Yixuan and Zhang, Zheng and Lin, Stephen and Guo, Baining},
  booktitle={Proceedings of the IEEE/CVF international conference on computer vision},
  pages={10012--10022},
  year={2021}
}

@inproceedings{dosovitskiy2020vit,
  title={An Image is Worth 16x16 Words: Transformers for Image Recognition at Scale},
  author={Dosovitskiy, Alexey and Beyer, Lucas and Kolesnikov, Alexander and Weissenborn, Dirk and Zhai, Xiaohua and Unterthiner, Thomas and Dehghani, Mostafa and Minderer, Matthias and Heigold, Georg and Gelly, Sylvain and others},
  booktitle={International Conference on Learning Representations},
  year={2021}
}


%% file: main.bib
@String(CVPR  = {IEEE Conf. Comput. Vis. Pattern Recog.})

@String(ICCV  = {Int. Conf. Comput. Vis.})

@String(ICLR  = {Int. Conf. Learn. Represent.})

@String(AAAI  = {AAAI})

@String(CVPR  = {CVPR})

@String(ICCV  = {ICCV})

@String(ICLR  = {ICLR})

@article{3DGS_2023,
  title    = {3D Gaussian Splatting for Real-Time Radiance Field Rendering},
  volume   = {42},
  issn     = {0730-0301, 1557-7368},
  doi      = {10.1145/3592433},
  number   = {4},
  journal  = {ACM Transactions on Graphics},
  author   = {Kerbl, Bernhard and Kopanas, Georgios and Leimkuehler, Thomas and Drettakis, George},
  year     = {2023},
  month    = aug,
  pages    = {1–14},
  language = {en}
}

@article{mildenhall2021nerf,
  title     = {Nerf: Representing scenes as neural radiance fields for view synthesis},
  author    = {Mildenhall, Ben and Srinivasan, Pratul P and Tancik, Matthew and Barron, Jonathan T and Ramamoorthi, Ravi and Ng, Ren},
  journal   = {Communications of the ACM},
  volume    = {65},
  number    = {1},
  pages     = {99--106},
  year      = {2021},
  publisher = {ACM New York, NY, USA}
}

@article{barron2023zipnerf,
  title   = {Zip-NeRF: Anti-Aliased Grid-Based Neural Radiance Fields},
  author  = {Jonathan T. Barron and Ben Mildenhall and
             Dor Verbin and Pratul P. Srinivasan and Peter Hedman},
  journal = {ICCV},
  year    = {2023}
}

@inproceedings{yeshwanthliu2023scannetpp,
  title={ScanNet++: A High-Fidelity Dataset of 3D Indoor Scenes},
  author={Yeshwanth, Chandan and Liu, Yueh-Cheng and Nie{\ss}ner, Matthias and Dai, Angela},
  booktitle = {Proceedings of the International Conference on Computer Vision ({ICCV})},
  year={2023}
}

@ARTICLE{Wang2004SSIM,
  author={Zhou Wang and Bovik, A.C. and Sheikh, H.R. and Simoncelli, E.P.},
  journal={IEEE Transactions on Image Processing}, 
  title={Image quality assessment: from error visibility to structural similarity}, 
  year={2004},
  volume={13},
  number={4},
  pages={600-612},
  doi={10.1109/TIP.2003.819861}
}

@inproceedings{zhang2018lpips,
  title={The Unreasonable Effectiveness of Deep Features as a Perceptual Metric},
  author={Zhang, Richard and Isola, Phillip and Efros, Alexei A and Shechtman, Eli and Wang, Oliver},
  booktitle={CVPR},
  year={2018}
}

@inproceedings{cao2025panda,
  title={Panda: Towards panoramic depth anything with unlabeled panoramas and mobius spatial augmentation},
  author={Cao, Zidong and Zhu, Jinjing and Zhang, Weiming and Ai, Hao and Bai, Haotian and Zhao, Hengshuang and Wang, Lin},
  booktitle={Proceedings of the IEEE/CVF Conference on Computer Vision and Pattern Recognition},
  pages={982--992},
  year={2025}
}

@inproceedings{guo2025depth,
  title={Depth any camera: Zero-shot metric depth estimation from any camera},
  author={Guo, Yuliang and Garg, Sparsh and Miangoleh, S Mahdi H and Huang, Xinyu and Ren, Liu},
  booktitle={Proceedings of the Computer Vision and Pattern Recognition Conference},
  pages={26996--27006},
  year={2025}
}

@inproceedings{piccinelli2025unik3d,
  title={Unik3d: Universal camera monocular 3d estimation},
  author={Piccinelli, Luigi and Sakaridis, Christos and Segu, Mattia and Yang, Yung-Hsu and Li, Siyuan and Abbeloos, Wim and Van Gool, Luc},
  booktitle={Proceedings of the Computer Vision and Pattern Recognition Conference},
  pages={1028--1039},
  year={2025}
}

@article{guo2026panovggt,
  title={PanoVGGT: Feed-Forward 3D Reconstruction from Panoramic Imagery},
  author={Guo, Yijing and Chao, Mengjun and Wang, Luo and Zhao, Tianyang and Dai, Haizhao and Zhang, Yingliang and Yu, Jingyi and Shi, Yujiao},
  journal={arXiv preprint arXiv:2603.17571},
  year={2026}
}

@article{lee2024odgs,
  title={Odgs: 3d scene reconstruction from omnidirectional images with 3d gaussian splattings},
  author={Lee, Suyoung and Chung, Jaeyoung and Huh, Jaeyoo and Lee, Kyoung Mu},
  journal={Advances in Neural Information Processing Systems},
  volume={37},
  pages={57050--57075},
  year={2024}
}

@inproceedings{li2025omnigs,
  title={Omnigs: Fast radiance field reconstruction using omnidirectional gaussian splatting},
  author={Li, Longwei and Huang, Huajian and Yeung, Sai-Kit and Cheng, Hui},
  booktitle={2025 IEEE/CVF Winter Conference on Applications of Computer Vision (WACV)},
  pages={2260--2268},
  year={2025},
  organization={IEEE}
}

@article{liao2024fisheye,
  title={Fisheye-gs: Lightweight and extensible gaussian splatting module for fisheye cameras},
  author={Liao, Zimu and Chen, Siyan and Fu, Rong and Wang, Yi and Su, Zhongling and Luo, Hao and Ma, Li and Xu, Linning and Dai, Bo and Li, Hengjie and others},
  journal={arXiv preprint arXiv:2409.04751},
  year={2024}
}

@inproceedings{wu20253dgut,
  title={3dgut: Enabling distorted cameras and secondary rays in gaussian splatting},
  author={Wu, Qi and Esturo, Janick Martinez and Mirzaei, Ashkan and Moenne-Loccoz, Nicolas and Gojcic, Zan},
  booktitle={Proceedings of the Computer Vision and Pattern Recognition Conference},
  pages={26036--26046},
  year={2025}
}

@inproceedings{wang2024360dvd,
  title={360dvd: Controllable panorama video generation with 360-degree video diffusion model},
  author={Wang, Qian and Li, Weiqi and Mou, Chong and Cheng, Xinhua and Zhang, Jian},
  booktitle={Proceedings of the IEEE/CVF Conference on Computer Vision and Pattern Recognition},
  pages={6913--6923},
  year={2024}
}

@inproceedings{zhang2026panflow,
  title={PanFlow: Decoupled Motion Control for Panoramic Video Generation},
  author={Zhang, Cheng and Liang, Hanwen and Chen, Donny Y and Wu, Qianyi and Plataniotis, Konstantinos N and Gambardella, Camilo Cruz and Cai, Jianfei},
  booktitle={Proceedings of the AAAI Conference on Artificial Intelligence},
  volume={40},
  number={15},
  pages={12385--12393},
  year={2026}
}

@article{tan2024imagine360,
  title={Imagine360: Immersive 360 video generation from perspective anchor},
  author={Tan, Jing and Yang, Shuai and Wu, Tong and He, Jingwen and Guo, Yuwei and Liu, Ziwei and Lin, Dahua},
  journal={arXiv preprint arXiv:2412.03552},
  year={2024}
}

@article{luo2025beyond,
  title={Beyond the frame: Generating 360 panoramic videos from perspective videos},
  author={Luo, Rundong and Wallingford, Matthew and Farhadi, Ali and Snavely, Noah and Ma, Wei-Chiu},
  journal={arXiv e-prints},
  pages={arXiv--2504},
  year={2025}
}

@article{yang2025matrix,
  title={Matrix-3d: Omnidirectional explorable 3d world generation},
  author={Yang, Zhongqi and Ge, Wenhang and Li, Yuqi and Chen, Jiaqi and Li, Haoyuan and An, Mengyin and Kang, Fei and Xue, Hua and Xu, Baixin and Yin, Yuyang and others},
  journal={arXiv preprint arXiv:2508.08086},
  year={2025}
}

@article{wang2025evoworld,
  title={Evoworld: Evolving panoramic world generation with explicit 3d memory},
  author={Wang, Jiahao and Ye, Luoxin and Lu, TaiMing and Xiao, Junfei and Zhang, Jiahan and Guo, Yuxiang and Liu, Xijun and Chellappa, Rama and Peng, Cheng and Yuille, Alan and others},
  journal={arXiv preprint arXiv:2510.01183},
  year={2025}
}

@article{team2025hunyuanworld,
  title={Hunyuanworld 1.0: Generating immersive, explorable, and interactive 3d worlds from words or pixels},
  author={Team, HunyuanWorld and Wang, Zhenwei and Liu, Yuhao and Wu, Junta and Gu, Zixiao and Wang, Haoyuan and Zuo, Xuhui and Huang, Tianyu and Li, Wenhuan and Zhang, Sheng and others},
  journal={arXiv preprint arXiv:2507.21809},
  year={2025}
}

@article{chen2024splatter360,
  title={Splatter-360: Generalizable 360$^\circ$ Gaussian Splatting for Wide-baseline Panoramic Images},
  author={Chen, Zheng and Wu, Chenming and Shen, Zhelun and Zhao, Chen and Ye, Weicai and Feng, Haocheng and Ding, Errui and Zhang, Song-Hai},
  journal={arXiv preprint arXiv:2412.06250},
  year={2024}
}
